\documentclass[letterpaper,twocolumn,10pt]{article}
\usepackage{usenix,epsfig,endnotes}

\usepackage{makecell,array,multirow}
\newcommand{\negs}{\vspace{-0.7\baselineskip}}
\usepackage[colorinlistoftodos,prependcaption,textsize=footnotesize,color=green]{todonotes}
\makeatletter
\usepackage{regexpatch}
\xpatchcmd{\@todo}{\setkeys{todonotes}{#1}}{\setkeys{todonotes}{inline,#1}}{}{}
\makeatother

\usepackage{xspace}
\usepackage{color}
\usepackage{csquotes}

\usepackage{amsmath}
\usepackage[margin=0.1cm,font=small]{caption}
\usepackage{subcaption}

\usepackage{booktabs}
\usepackage{mdwtab}
\usepackage{soul}
\usepackage{bbm}
\usepackage{tabularx}
\usepackage{hhline}

\usepackage{enumitem}

\newcommand{\pribot}{{\fontfamily{cmss}\selectfont{PriBot}}\xspace}
\newcommand{\framework}{{\fontfamily{cmss}\selectfont{Polisis}}\xspace}

\newcommand*{\new}{\textcolor{black}}
\newcommand{\qa}{QA\xspace}

\newcommand{\random}{{\fontfamily{cmss}\selectfont{Random}}\xspace}
\newcommand{\retrieval}{{\fontfamily{cmss}\selectfont{Retrieval}}\xspace}
\newcommand{\etoe}{{\fontfamily{cmss}\selectfont{SemVec}}\xspace}

\newcommand{\wpEmb}{{\fontfamily{cmss}\selectfont{WP}}\xspace}
\newcommand{\wpNoSubEmb}{{\fontfamily{cmss}\selectfont{WP-NoSub}}\xspace}
\newcommand{\peEmb}{{\fontfamily{cmss}\selectfont{PE}}\xspace}
\newcommand{\peNoSubEmb}{{\fontfamily{cmss}\selectfont{PE-NoSub}}\xspace}

\newcommand{\siteurl}{\href{https://pribot.org/}{\url{https://pribot.org}}\xspace}

\definecolor{lightpink}{HTML}{FFB6C1}
\definecolor{orchid}{HTML}{DA70D6}
\definecolor{darkorchid}{HTML}{9932CC}
\definecolor{midnightblue}{HTML}{191970}

\definecolor{green1}{HTML}{EDF8FB}
\definecolor{green2}{HTML}{B2E2E2}
\definecolor{green3}{HTML}{66C2A4}
\definecolor{green4}{HTML}{238B45}

\newcommand{\topk}{{\fontfamily{cmss}\selectfont{top-$k$ score}}\xspace}

\newcommand{\MAP}{{\fontfamily{cmss}\selectfont{MAP}}\xspace}

\newcommand{\NDCG}{{\fontfamily{cmss}\selectfont{NDCG}}\xspace}

\newcolumntype{C}[1]{>{\centering\let\newline\\\arraybackslash\hspace{0pt}}m{#1}}
\newcolumntype{L}[1]{>{\raggedright\let\newline\\\arraybackslash\hspace{0pt}}m{#1}}
\usepackage{bm}

\usepackage{url}

\usepackage{breakurl}

\usepackage[hidelinks,breaklinks]{hyperref}
\hypersetup{
	colorlinks,
	linkcolor={black!50!black},
	citecolor={black!50!black},
	urlcolor={black!80!black}
}

\usepackage{mathtools}
\usepackage{titlesec}
\titlespacing\section{0pt}{5pt plus 4pt minus 2pt}{3pt plus 2pt minus 2pt}
\titlespacing\subsection{0pt}{5pt plus 4pt minus 2pt}{3pt plus 2pt minus 2pt}
\titlespacing\subsubsection{0pt}{5pt plus 4pt minus 2pt}{3pt plus 2pt minus 2pt}
\titlespacing\paragraph{0pt}{2pt plus 4pt minus 2pt}{4pt plus 2pt minus 2pt}

\newcommand\kmmm[1]{\textcolor{black}{#1}}
\newcommand\hmmm[1]{\textcolor{black}{#1}}
\newcommand\AppendixCaption[1]{%
	}

\begin{document}
\date{}

\title{\Large \bf Polisis: Automated Analysis and\\Presentation of Privacy Policies Using Deep Learning}

\author{
{\rm Hamza Harkous\textsuperscript{1},  Kassem Fawaz\textsuperscript{2}, R\'emi Lebret\textsuperscript{1}, Florian Schaub\textsuperscript{3}, Kang G. Shin\textsuperscript{3}, and Karl Aberer\textsuperscript{1}}
\\
\\
\rm \textsuperscript{1} \'Ecole Polytechnique F\'ed\'erale de Lausanne (EPFL)\\
\rm \textsuperscript{2} University of of Wisconsin-Madison\\
\rm \textsuperscript{3} University of Michigan
}
\maketitle
\begin{abstract}
Privacy policies are the primary channel through which companies inform users about their data collection and sharing practices. These policies are often long and difficult to comprehend. 
Short notices based on information extracted from privacy policies have been shown to be useful but face a significant \textit{scalability} hurdle, given the number of policies and their evolution over time. 
Companies, users, researchers, and regulators still lack usable and scalable tools to cope with the breadth and depth of privacy policies.
To address these hurdles, we propose an automated framework for privacy \underline{\bf {poli}}cy analy\underline{\bf {sis}} ({\bf\framework}). It enables scalable, dynamic, and multi-dimensional queries on natural language privacy policies. At the core of \framework is a privacy-centric language model, built with 130K privacy policies, and a novel hierarchy of neural-network classifiers that accounts for both high-level aspects and fine-grained details of privacy practices. We demonstrate \framework' modularity and utility with two applications  supporting \textit{structured} and \textit{free-form} querying. The structured querying application is the automated assignment of privacy icons from privacy policies. With \framework, we can achieve an accuracy of 88.4\% on this task. The second application, \pribot, is the first free-form question-answering system for privacy policies. We show that \pribot can produce a correct answer among its top-3 results for 82\% of the test questions. Using an MTurk user study with 700 participants, we show that at least one of \pribot's top-3 answers is relevant to users for 89\% of the test questions.
\end{abstract}

\section{Introduction}
\label{sec:introduction}

Privacy policies are one of the most common ways of providing notice and choice online.
They aim to inform users how companies collect, store and manage their personal information. Although some service providers have improved the comprehensibility and readability of their privacy policies, these policies remain excessively long and difficult to follow~\cite{cate:2010,ftc:2012,Gluck:2016,mcdonald2008cost,wh:2014}. 
In 2008, McDonald and Cranor~\cite{mcdonald2008cost} estimated that it would take an average user 201 hours to read all the privacy policies encountered in a year. Since then, we have witnessed a smartphone revolution and the rise of the Internet of Things (IoTs), which lead to the proliferation of services and associated policies~\cite{schaub2017}.
In addition, emerging technologies brought along new forms of user interfaces (UIs), such as voice-controlled devices or wearables, for which existing techniques for presenting privacy policies are not suitable~\cite{Gluck:2016,schaub2017,ftc:2015,Schaub:2015}.

\paragraph*{\textbf{Problem Description.}}
Users, researchers, and regulators are not well-equipped to process or understand the content of privacy policies, especially at scale. Users are surprised by data practices that do not meet their expectations~\cite{Rao:2017:soups}, hidden in long, vague, and ambiguous policies. Researchers employ expert annotators to analyze and reason about a subset of the available privacy policies~\cite{Wilson:2016,Wilsonacl16}. Regulators, such as the U.S. Department of Commerce, rely on companies to self-certify their compliance with privacy practices (e.g., the Privacy Shield Framework~\cite{privacyshield}). The \textit{problem} lies in stakeholders lacking the usable and scalable tools to deal with the breadth and depth of privacy policies.

Several proposals have aimed at alternative methods and UIs for presenting privacy notices~\cite{Schaub:2015}, including machine-readable formats~\cite{cranor2002web}, nutrition labels~\cite{Kelley:2009:NLP:1572532.1572538}, privacy icons (recently recommended by the EU~\cite{eu:gdpr}), and short notices~\cite{zimmeck2014privee}. Unfortunately, these approaches have faced a significant \textit{scalability} hurdle: the human effort needed to retrofit the new notices to existing policies and maintain them over time is tremendous. The existing research towards automating this process has been limited in scope to a handful of ``queries,'' e.g., whether the policy mentions data encryption or whether it provides an opt-out choice from third-party tracking ~\cite{zimmeck2014privee,sathyendra2017identifying}.

\paragraph*{\textbf{Our Framework.}}

We overcome this scalability hurdle by proposing an automatic and comprehensive framework for privacy \underline{{\bf poli}}cy analy\underline{{\bf sis}}  ({\bf\framework}). It divides a privacy policy into smaller and self-contained fragments of text, referred to as {\em segments}. \framework automatically annotates, with high accuracy, each segment with a set of labels describing its data practices.   
Unlike prior research in automatic labeling/analysis of privacy policies, \framework does not just predict a handful of classes given the entire policy document. Instead, \framework annotates the privacy policy at a much finer-grained scale. It predicts for each segment the set of classes that account for both the high-level aspects and the fine-grained classes of embedded privacy information. \framework uses these classes to enable scalable, dynamic, and multi-dimensional queries on privacy policies, in a way not possible with prior approaches. 

At the core of \framework is a novel hierarchy of neural-network classifiers that involve 10 high-level and 122 fine-grained privacy classes for privacy-policy segments. To build these fine-grained classifiers, we leverage techniques such as subword embeddings and multi-label classification. We further seed these classifiers with a custom, privacy-specific language model that we generated using our corpus of more than 130,000 privacy policies from websites and mobile apps. 

\framework provides the underlying intelligence for researchers and regulators to focus their efforts on merely designing a set of queries that power their applications. We stress, however, that \framework is not intended to replace the privacy policy -- as a legal document -- with an automated interpretation. Similar to existing approaches on privacy policies' analysis and presentation, it decouples the legally binding functionality of these policies from their informational utility.

\paragraph*{\textbf{Applications.}}

We demonstrate and evaluate the modularity and utility of \framework\ with two robust applications that support \textit{structured} and \textit{free-form} querying of privacy policies. 

The \textit{structured querying} application involves extracting short notices in the form of privacy icons from privacy policies. As a case study, we investigate the Disconnect privacy icons~\cite{disconnect_icons}. 
By composing a set of simple rules on top of \framework, we show a solution that can automatically select appropriate privacy icons from a privacy policy. We further study the practice of companies assigning icons to privacy policies at scale. We empirically demonstrate that existing privacy-compliance companies, such as TRUSTe (now rebranded as TrustArc), might be adopting permissive policies when assigning such privacy icons. Our findings are consistent with anecdotal controversies and manually investigated issues in privacy certification and compliance processes~\cite{Edelman:2009,trustefb,miyazaki2002internet}.

The second application illustrates the power of \textit{free-form querying} in \framework. We design, implement and evaluate \pribot, the first automated Question-Answering ({\qa}) system for privacy policies. \pribot extracts the relevant privacy policy segments to answer the user's free-form questions. To build \pribot, we overcame the non-existence of a public, privacy-specific \qa dataset by casting the problem as a ranking problem that could be solved using the classification results of \framework. \pribot matches user questions with answers from a previously unseen privacy policy, in real time and with high accuracy -- demonstrating a more intuitive and user-friendly way to present privacy notices and controls. We evaluate \pribot using a new test dataset, based on real-world questions that have been asked by consumers on Twitter.

\paragraph*{\textbf{Contributions.}}
With this paper we make the following contributions:
\begin{itemize}[leftmargin=*]
\item We design and implement \framework, an approach for automatically annotating previously unseen privacy policies with high-level and fine-grained labels from a pre-specified taxonomy (Sec.~\ref{sec:framework},~\ref{sec:data_layer},~\ref{sec:ml_layer}, and~\ref{sec:app_layer}). 
\item We demonstrate how \framework can be used to assign privacy icons to a privacy policy with an average accuracy of 88.4\%. This accuracy is computed by comparing icons assigned with \framework' automatic labels to icons assigned based on manual annotations by three legal experts from the OPP-115 dataset~\cite{Wilsonacl16} (Sec.~\ref{sec:privacy_icons}).
\item We design, implement and evaluate \pribot, a \qa system that answers free-form user questions from privacy policies (Sec.~\ref{sec:approaches}). Our accuracy evaluation shows that \pribot produces at least one correct answer (as indicated by privacy experts) in its top three for 82\% of the test questions and as the top one for 68\% of the test questions. Our evaluation of the perceived utility with 700 MTurk crowdworkers shows that users find a relevant answer in \pribot's top-3 for 89\% of the questions (Sec.~\ref{sec:qa_evaluation}).

\item We make \framework publicly available by providing three web services demonstrating our applications: a service giving a visual overview of the different aspects of each privacy policy, a chatbot for answering user questions in real time, and a privacy-labels interface for privacy policies. These services are available at \siteurl. We provide screenshots of these applications in Appendix B.
\end{itemize}

\section{Framework Overview}
\label{sec:framework}

\begin{figure}[t]
\centering
	\includegraphics[width=1\linewidth]{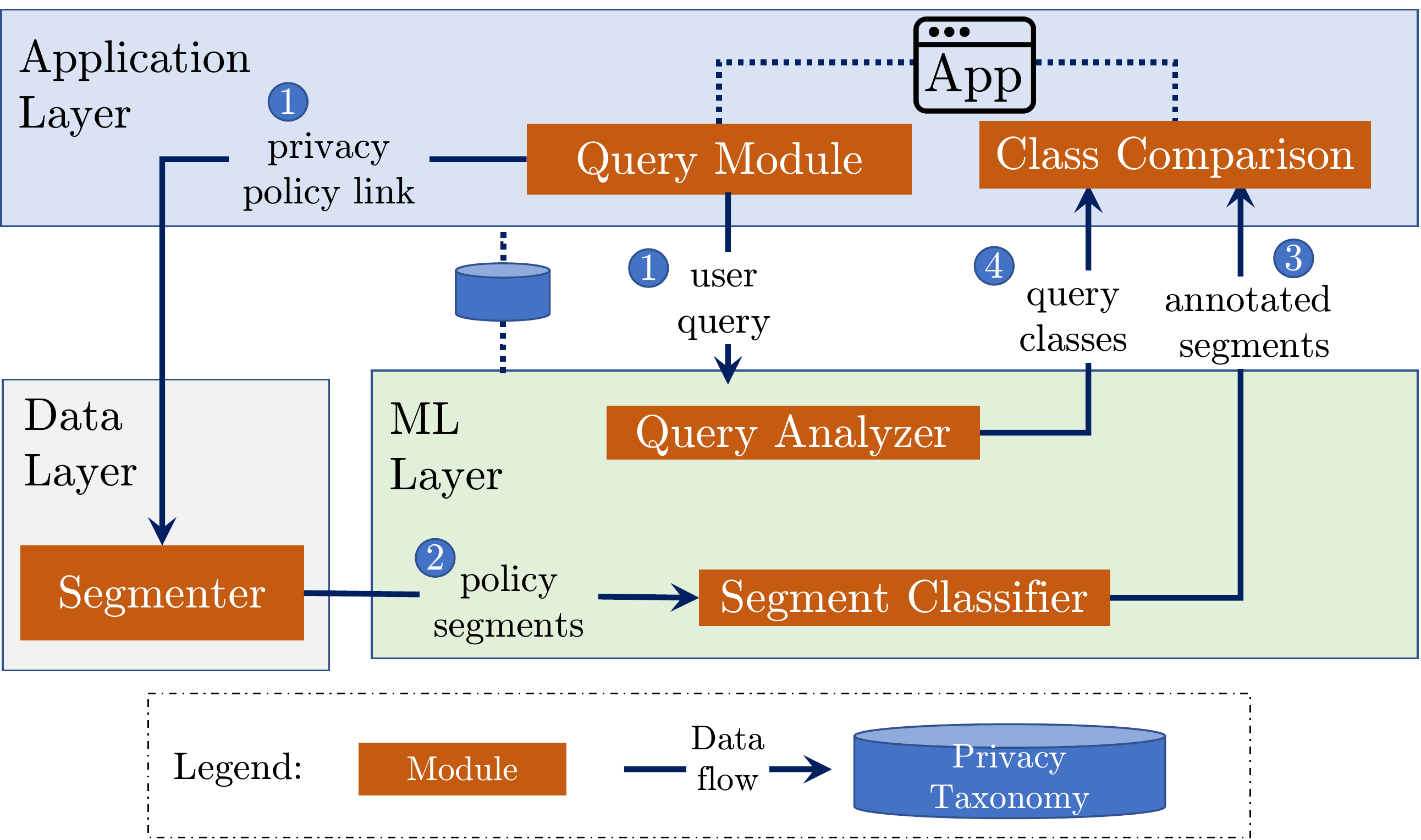}
	\caption{A high-level overview of \framework.}
	\label{fig:framework}
\end{figure}

Fig.~\ref{fig:framework} shows a high-level overview of \framework. It comprises three layers:  \textit{Application Layer}, \textit{Data Layer}, and \textit{Machine Learning (ML) Layer}. \framework treats a privacy policy as a list of semantically coherent segments (i.e., groups of consecutive sentences). It also utilizes a taxonomy of privacy data practices. One example of such a taxonomy was introduced by Wilson {\em et al.}~\cite{Wilsonacl16} (see also Fig.~\ref{fig:opp} in Sec.~\ref{sec:ml_layer}).

\textbf{Application Layer} (Sec.~\ref{sec:app_layer}, ~\ref{sec:privacy_icons} \&~\ref{sec:approaches}):
The Application Layer provides fine-grained information about the privacy policy, thus providing the users with high modularity in posing their queries. In this layer, a \textit{Query Module} receives the \textit{User Query} about a privacy policy (Step 1 in Fig.~\ref{fig:framework}). These inputs are forwarded to lower layers, which then extract the privacy classes embedded within the query and the policy's segments. To resolve the user query, the \textit{Class-Comparison} module identifies the segments with privacy classes matching those of the query. Then, it passes the matched segments (with their predicted classes) back to the application. 

  \textbf{Data Layer} (Sec.~\ref{sec:data_layer}): The Data Layer first scrapes the policy's webpage. Then, it partitions the policy into semantically coherent and adequately sized segments (using the \textit{Segmenter} component in Step 2 of Fig.~\ref{fig:framework}). Each of the resulting segments can be independently consumed by both the humans and programming interfaces.
  
\textbf{Machine Learning Layer} (Sec.~\ref{sec:ml_layer}): 
In order to enable a multitude of applications to be built around \framework, the ML layer is responsible for producing rich and fine-grained annotations of the data segments. 
This layer takes as an input the privacy-policy segments from the Data Layer (Step 2) and the user query (Step 1) from the Application Layer. The \textit{Segment Classifier} probabilistically assigns each segment a set of class--value pairs describing its data practices. For example, an element in this set can be \textsl{information-type=location} with probability $p=0.65$.
Similarly, the \textit{Query Analyzer} extracts the privacy classes from the user's query.  Finally, the class--value pairs of both the segments and the query are passed back to the Class Comparison module of the Application Layer (Steps 3 and 4).

\section{Data Layer}
\label{sec:data_layer}
 \begin{figure}[t]
  \includegraphics[width=\linewidth]{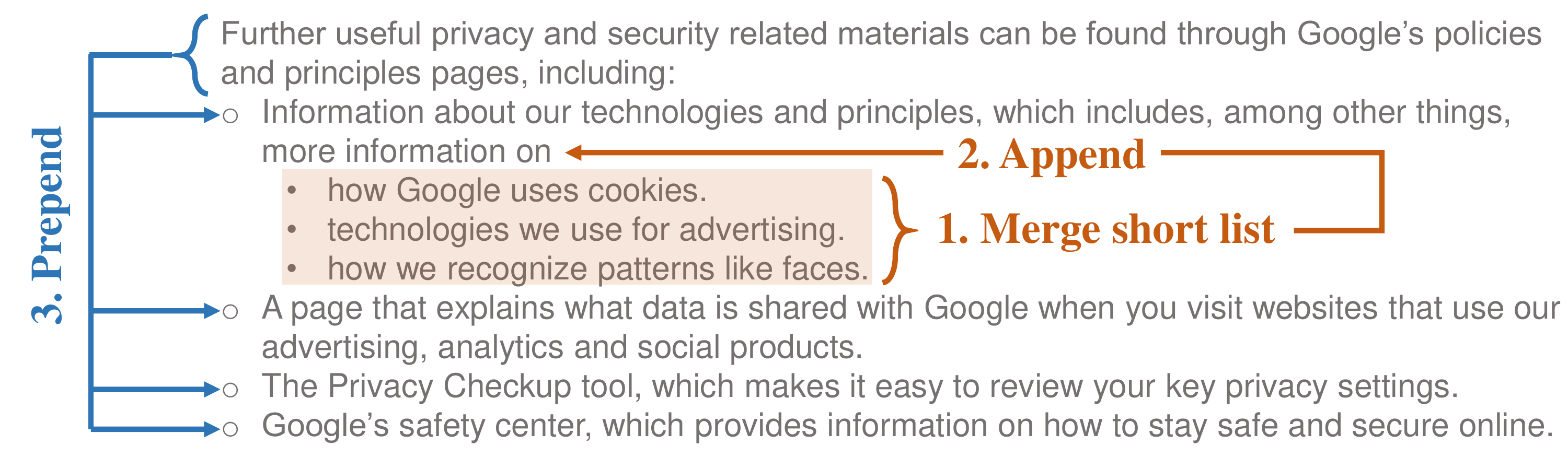}
  \caption{List merging during the policy segmentation.}
  \label{fig:segment_example}
\end{figure}

To pre-process the privacy policy, the Data Layer employs a \textit{Segmenter} module in three stages: extraction, list handling, and segmentation. The Data Layer requires no information other than the link to the privacy policy.

\paragraph*{Policy Extraction:}
Given the URL of a privacy policy, the segmenter employs Google Chrome in headless mode (without UI) to scrape the policy's webpage. It waits for the page to fully load which happens after all the JavaScript has been downloaded and executed.  Then, the segmenter removes all irrelevant HTML elements including the scripts, header, footer, side/navigation menus, comments, and CSS.

Although several online privacy policies contain dynamically viewable content (e.g., accordion toggles and collapsible/expandable paragraphs), the ``dynamic" content is already part of the loaded webpage in almost all cases. For example, when the user expands a collapsible paragraph, a local JavaScript exposes an offline HTML snippet; no further downloading takes place. 

We confirmed this with the privacy policies of the top 200 global websites from Alexa.com.
For each privacy-policy link, we compared the segmenter's scraped content to that extracted from our manual navigation of the same policy (while accounting for all the dynamically viewable elements of the webpage). 
Using a fuzzy string matching library,\footnote{\url{https://pypi.python.org/pypi/fuzzywuzzy}} we found that the segmenter's scraped policy covers, on average, 99.08\% of the content of the manually fetched policy.

\paragraph{List Aggregation:}
Second, the segmenter handles any ordered/unordered lists inside the policy. Lists require a special treatment since counting an entire lengthy list, possibly covering diverse data practices, as a single segment could result in noisy annotations. On the other hand, treating each list item as an independent segment is problematic as list elements are typically not self-contained, resulting in missed annotations. 
See Fig.~\ref{fig:segment_example} from Google's privacy policy as an example\footnote{\url{https://www.google.com/intl/en_US/policies/privacy/archive/20160829/}, last modified on Aug.~29, 2016, retrieved on Jun.~27, 2018}. 

Our handling of the lists involves two techniques: one for short list items (e.g., the inner list of Fig.~\ref{fig:segment_example}) and another for longer list items (e.g., the outer list of Fig.~\ref{fig:segment_example}). 
For short list items (maximum of 20 words per element), the segmenter combines the elements with the introductory statement of the list into a single paragraph element (with \texttt{<p>} tag). The rest of the lists with long items are transformed into a set of paragraphs. Each paragraph is a distinct list element prepended by the list's introductory statement (Step 3 in Fig.~\ref{fig:segment_example}).

\paragraph{Policy Segmentation:}
The segmenter performs an initial coarse segmentation by breaking down the policy according to the HTML \texttt{<div>} and \texttt{<p>} tags. 
The output of this step is an initial set of policy segments. 
As some of the resulting segments might still be long, we subdivide them further with another technique. We use \texttt{GraphSeg}~\cite{graphseg}, an unsupervised algorithm that generates semantically coherent segments. It relies on word embeddings to generate segments as cliques of related (semantically similar) sentences. For that purpose, we use custom, domain-specific word embeddings that we generated using our corpus of 130K privacy policies (\textit{cf.} Sec.~\ref{sec:ml_layer}). Finally, the segmenter outputs a series of fine-grained segments to the Machine Learning Layer, where they are automatically analyzed.

\section{Machine Learning Layer}
\label{sec:ml_layer}
This section describes the components of \framework' Machine Learning Layer in two stages: (1) an \textit{unsupervised} stage, in which we build domain-specific word vectors (i.e., word embeddings) for privacy policies from unlabeled data, and (2) a \textit{supervised stage}, in which we train a novel hierarchy of privacy-text classifiers, based on neural networks, that leverages the word vectors. These classifiers power the \textit{Segment Classifier} and \textit{Query Analyzer} modules of Fig.~\ref{fig:framework}.
We use word embeddings and neural networks thanks to their proven advantages in text classification~\cite{Kim14} over traditional techniques. 

\subsection{Privacy-Specific Word Embeddings}

Traditional text classifiers use the words and their frequencies as the building block for their features. They, however, have limited generalization power, especially when the training datasets are limited in size and scope.  For example, replacing the word ``erase'' by the word ``delete'' can significantly change the classification result if ``delete'' was not in the classifier's training set.

Word embeddings solve this issue by extracting generic word vectors from a large corpus, in an unsupervised manner, and enabling their use in new classification problems (a technique termed \textit{Transfer Learning}). The features in the classifiers become the word vectors instead of the words themselves. Hence, two text segments composed of semantically similar words would be represented by two groups of word vectors (i.e., features) that are close in the vector space. This allows the text classifier to account for words outside the training set, as long as they are part of the large corpus used to train the word vectors.

While general-purpose pre-trained embeddings, such as Word2vec~\cite{mikolov2013distributed} and GloVe~\cite{pennington2014glove} do exist, domain-specific embeddings %
result in better classification accuracy~\cite{tang2014learning}. Thus, we trained custom word embeddings for the privacy-policy domain. To that end, we created a corpus of 130K privacy policies collected from apps on the Google Play Store. These policies typically describe the overall data practices of the apps' companies.

We crawled the metadata of more than 1.4 million Android apps available via the PlayDrone project~\cite{viennot2014measurement} to find the links to 199,186 privacy policies. We crawled the web pages for these policies, retrieving 130,326 policies which returned an HTTP status code of 200. Then, we extracted the textual content from their HTML using the policy crawler described in Sec.~\ref{sec:data_layer}. 
We will refer to this corpus as the \textit{Policies Corpus}. 
Using this corpus, we trained a word-embeddings model using \textit{fastText}~\cite{bojanowski2016enriching}. We henceforth call this model the \textit{Policies Embeddings}. A major advantage of using \textit{fastText} is that it allows training vectors for \textit{subwords} (or character $n$-grams of sizes 3 to 6) in addition to words. Hence, even if we have words outside our corpus, we can assign them vectors by combining the vectors of their constituent subwords. This is very useful in accounting for spelling mistakes that occur in applications that involve free-form user queries.

\subsection{Classification Dataset}
\label{sec:data_over}
\begin{figure*}[t]
	\centering
  \includegraphics[width=1\textwidth]{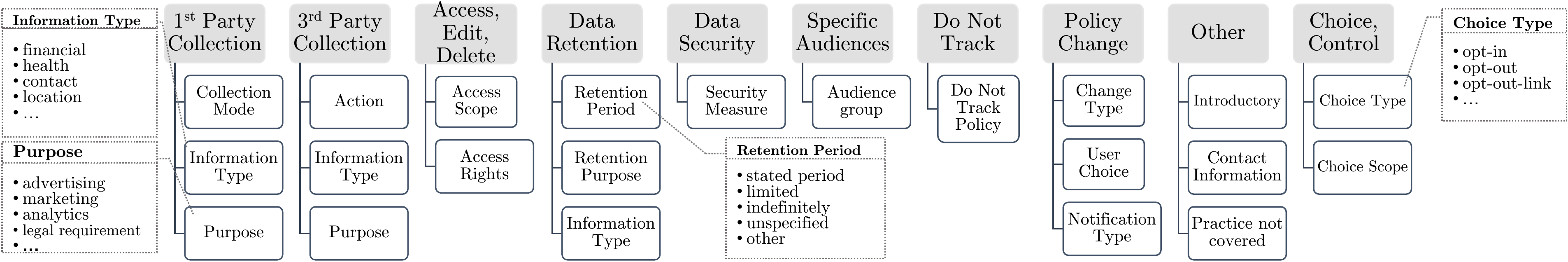}
	\caption{ The privacy taxonomy of Wilson {\em et al.}~\cite{Wilsonacl16}. The top level of the hierarchy (shaded blocks) defines high-level privacy categories. The lower level defines a set of privacy attributes, each assuming a set of values. We show examples of values for some of the attributes.}
	\label{fig:opp}
\end{figure*}

Our Policies Embeddings provides a solid starting point to build robust classifiers. However, training the classifiers to detect fine-grained labels of privacy policies' segments requires a labeled dataset. For that purpose, we leverage the \textit{Online Privacy Policies} (OPP-115) dataset, introduced by Wilson \textit{et al.}~\cite{Wilsonacl16}.
This dataset contains 115 privacy policies manually annotated by skilled annotators (law school students). In total, the dataset has 23K annotated data practices. The annotations were at two levels. First, paragraph-sized segments were annotated according to one or more of the 10 high-level categories in Fig.~\ref{fig:opp} (e.g., \textsl{First Party Collection}, \textsl{Data Retention}). Then, annotators selected parts of the segment and annotated them using attribute--value pairs,  e.g., \textsl{information\_type: location}, \textsl{purpose: advertising}, etc.
In total, there were 20 distinct attributes and 138 distinct values across all attributes. Of these, 122 values had more than 20 labels. In Fig.~\ref{fig:opp}, we only show the mandatory attributes that should be present in all segments. Due to space limitation, we only show samples of the values for selected attributes in Fig.~\ref{fig:opp}.

\subsection{Hierarchical Multi-label Classification}

To account for the multiple granularity levels in the policies' text, we build a hierarchy of classifiers that are individually trained on handling specific parts of the problem. 

At the \textbf{top level}, a classifier predicts one or more high-level categories of the input segment $x$ (categories are the top-level, shaded boxes of Fig.~\ref{fig:opp}). We train a multi-label classifier that provides us with the probability $p(c_i|x)$ of the occurrence of each high-level category $c_i$, taken from the set of all categories $\mathcal{C}$. In addition to allowing multiple categories per segment, using a multi-label classifier makes it possible to determine whether a category is present in a segment by simply comparing its classification probability to a threshold of 0.5.

At the \textbf{lower level}, a set of classifiers predicts one or more values for each privacy attribute (the leaves in the taxonomy of Fig.~\ref{fig:opp}).
We train a set of multi-label classifiers on the attribute-level. Each classifier produces the probabilities $p(v_j|x)$ for the values $v_j \in \mathcal{V}(b)$ of a single attribute $b$. For example, given the attribute \mbox{\textsl{b=information\_type}}, the corresponding classifier outputs the probabilities for elements in $\mathcal{V}(b)$: $\{$\textsl{financial, location, user profile, health, demographics, cookies, contact information, generic personal information, \mbox{unspecified, \ldots$\}$.}} 

An important consequence of this hierarchy is that interpreting the output of the attribute-level classifier depends on the categories' probabilities. For example, the values' probabilities of the attribute \textsl{``retention\_period''} are irrelevant when the dominant high-level category is \textsl{``policy\_change.''} Hence, for a category $c_i$, one would only consider the attributes descending from it in the hierarchy. We denote these attributes as $\mathcal{A}(c_i)$ and the set of all values across these attributes as $\mathcal{V}(c_i)$.

We use Convolutional Neural Networks (CNNs) internally within all the classifiers for two main reasons, \hmmm{which are also common in similar classification tasks.}
First, CNNs enable us to integrate pre-trained word embeddings that provide the classifiers with better generalization capabilities. Second, CNNs recognize when a certain set of tokens are a good indicator of the class, in a way that is invariant to their position within the input segment.

\begin{figure}[t]
	\centering
	\includegraphics[width=0.9\linewidth]{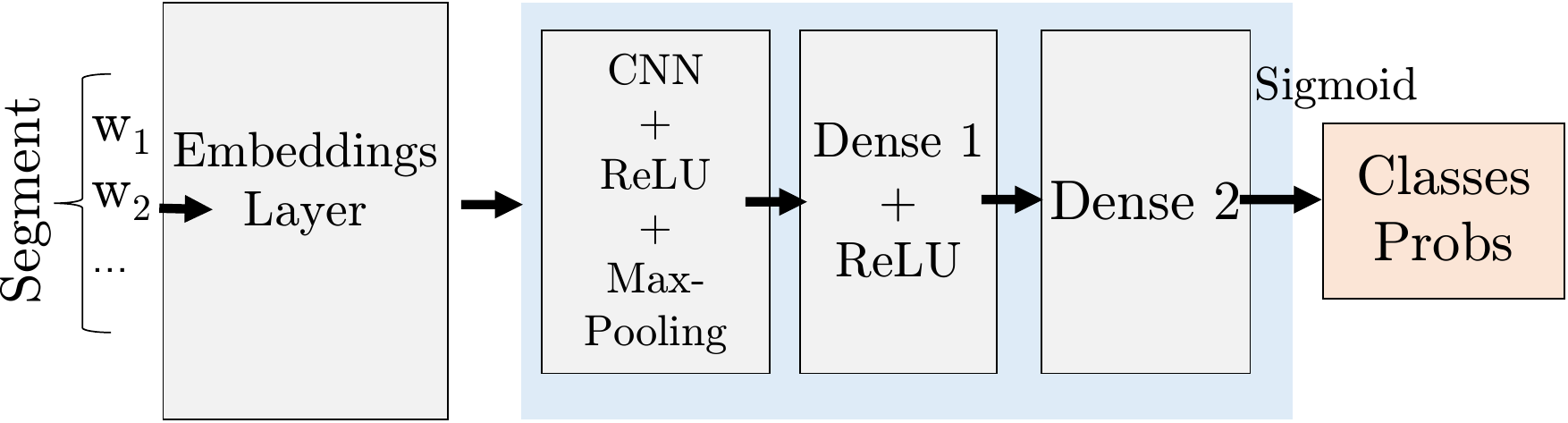}
	\captionof{figure}{Components of the CNN-based classifier used.}
	\label{fig:class_approach}

\end{figure}

We use a similar CNN architecture for classifiers on both levels as shown in Fig.~\ref{fig:class_approach}. 
\hmmm{Segments are split into tokens, using PENN Treebank tokenization in NLTK~\cite{bird2004nltk}. The embeddings layer outputs the word vectors of these tokens. We froze that layer, preventing its weights from being updated, in order to preserve the learnt semantic similarity between all the words present in our Policies Embeddings. Next, the word vectors pass through a Convolutional layer, whose main role is applying a non-linear function (a Rectified Linear Unit (ReLU)) over windows of $k$ words. Then, a max-pooling layer combines the vectors resulting from the different windows into a single vector. This vector then passes through the first dense (i.e., fully-connected) layer with a ReLU activation function, and finally through the second dense layer.} A \textit{sigmoid} operation is applied to the output of the last layer to obtain the probabilities for the possible output classes. We used \textit{multi-label cross-entropy loss} as the classifier's objective function. 
\hmmm{We refer interested readers to~\cite{britz2015understanding} for further elaborations on how CNNs are used in such contexts.}

\paragraph*{Models' Training.}
In total, we trained 20 classifiers at the attribute level (including the optional attributes). We also trained two classifiers at the category level: one for classifying segments and the other for classifying free-form queries. For the former, we include all the classes in Fig.~\ref{fig:opp}. For the latter, we ignore the \textsl{``Other''} category as it is mainly for introductory sentences or uncovered practices~\cite{Wilsonacl16}, which are not applicable to users' queries.
For training the classifiers, we used the data from 65 policies in the OPP-115 dataset, and we kept 50 policies as a testing set. The hyper-parameters for each classifier were obtained by running a randomized grid-search. 
In Table~\ref{table:class_cat}, we present the evaluation metrics on the testing set for the category classifier intended for free-form queries. In addition to the precision, recall and F1 scores (macro-averaged per label\footnote{A successful multilabel classifier should not only predict the \textit{presence} of a label, but also its \textit{absence}. Otherwise, a model that predicts that all labels are present would have 100\% precision and recall. For that, the precision in the table represents the macro-average of the precision in predicting the presence of each label and predicting its absence (similarly for recall and F1 metrics).}), we also show the top-1 precision metric, representing the fraction of segments where the top predicted category label occurs in the annotators' ground-truth labels. As evident in the table, our classifiers can predict the top-level privacy category with high accuracy. Although we consider the problem in the multi-label setting, these metrics are significantly higher than the models presented in the original OPP-115 paper~\cite{Wilsonacl16}. The full results for the rest of classifiers are presented in Appendix A. The efficacy of these classifiers is further highlighted through queries that directly leverage their output in the applications described next.
\begin{table}[t]
\vspace{0.5\baselineskip}
	\renewcommand{\arraystretch}{1.05}
	\scriptsize
	\centering
	\caption{Classification results for user queries at the category level. Hyperparameters: Embeddings size: 300, Number of filters: 200, Filter Size: 3, Dense Layer Size: 100, Batch Size: 40}
	\begin{tabular}{lccC{0.5cm} C{0.9cm} C{0.9cm}}
		\toprule
        \negs
		\textbf{Category}  &\textbf{Prec.}  &\textbf{Recall} & \textbf{F1}& \textbf{Top-1 Prec.} & \textbf{Support}\\
		\midrule
        $1^{\mbox{st}}$ Party Collection & 0.80 & 0.80 & 0.80 & 0.80 & 1267\\
         $3^{\mbox{rd}}$ Party Sharing & 0.81 & 0.81 & 0.81 & 0.86 & 963\\
         User Choice/Control & 0.76 & 0.73 & 0.75 & 0.81 & 455\\
    Data Security & 0.87 & 0.86 & 0.87 & 0.77 & 202\\
         Specific Audiences & 0.95 & 0.94 & 0.95 & 0.91 & 156\\
        Access, Edit, Delete & 0.94 & 0.75 & 0.82 & 0.97 & 134\\
         Policy Change & 0.96 & 0.89 & 0.92 & 0.93 & 120\\
Data Retention & 0.79 & 0.67 & 0.71 & 0.60 & 93\\
        Do Not Track  & 0.97 & 0.97 & 0.97 & 0.94 & 16\\
		\midrule
		Average  & {0.87}&  {0.83} & {0.84} & {0.84}\\
		\bottomrule
	\end{tabular}
	\label{table:class_cat}
\end{table}

\section{Application Layer}
\label{sec:app_layer}

Leveraging the power of the ML Layer's classifiers, \framework supports both \textit{structured} and \textit{free-from} queries about a privacy policy's content. A structured query is a combination of first-order logic predicates over the predicted privacy classes and the policy segments, such as: $\exists s \ (s \in \mbox{\textsl{policy}} \land \mbox{\textsl{information\_type(s)=location}} \land \mbox{\textsl{purpose(s)}} = \mbox{\textsl{marketing}}  \land \mbox{\textsl{user\_choice(s)=opt-out}})$. 
On the other hand, a free-form query is simply a natural language question posed directly by the users, such as {\fontfamily{cmss}\selectfont \small {``do you share my location with third parties?"}}. The response to a query is the set of segments satisfying the predicates in the case of a structured query or matching the user's question in the case of a free-form query. The Application Layer builds on these query types to enable an array of applications for different privacy stakeholders. We take an exemplification approach to give the reader a better intuition on these applications, before delving deeper into two of them in the next sections.

\paragraph*{Users:}
\framework can automatically populate several of the previously-proposed short notices for privacy policies, such as nutrition tables and privacy icons~\cite{Gluck:2016,disconnect_icons,cranor2006user,kelley2009nutrition}. This task can be achieved by mapping the notices to a set of structured queries (\textit{cf.} Sec.~\ref{sec:privacy_icons}).
Another possible application is privacy-centered comparative shopping~\cite{tsai2011effect}. A user can build on \framework' output to automatically quantify the privacy utility of a certain policy. For example, such a privacy metric could be a combination of positive scores describing privacy-protecting features (e.g., policy containing a segment with the label: \textsl{retention\_period: stated period}) and negative scores describing privacy-infringing features (e.g., policy containing a segment with the label: \textsl{retention\_period: unlimited}). A major advantage of automatically generating short notices is that they can be seamlessly refreshed when policies are updated or when the rules to generate these notices are modified. Otherwise, discrepancies between policies and notices might arise over time, which deters companies from adopting the short notices in the first place.

By answering free-form queries with relevant policy segments, \framework can remove the interface barrier between the policy and the users, especially in conversational interfaces (e.g., voice assistants and chatbots). Taking a step further, \framework' output can be potentially used to automatically rephrase the answer segments to a simpler language. A rule engine can generate text based on the combination of predicted classes of an answer segment (e.g., {\fontfamily{cmss}\selectfont \small {``We share data with third parties. This concerns our users' information, like your online activities. We need this to respond to requests from legal authorities"}}).

\textbf{Researchers:}
The difficultly of analyzing the data-collection claims by companies at scale has often been cited as a limitation in ecosystem studies (e.g.,~\cite{razaghpanahapps}). 
\framework can provide the means to overcome that. For instance, researchers interested in analyzing apps that admit collecting health data~\cite{aktypiunwinding,steel2013health} could utilize \framework to query a dataset of app policies. One example query can be formed by joining the label \textsl{information\_type: health} with the category of \textsl{First Party Collection} or \textsl{Third Party Sharing}. 

\textbf{Regulators:} Numerous studies from regulators and law and public policy researchers have manually analyzed the permissiveness of compliance checks ~\cite{miyazaki2002internet,appfail}. The number of assessed privacy policies in these studies is typically in the range of tens of policies. For instance, the Norwegian Consumer Council has investigated the level of ambiguity in defining personal information within only 20 privacy policies~\cite{appfail}. \framework can scale such studies by processing a regulator's queries on large datasets. For example, with \framework, policies can be ranked according to an automated ambiguity metric by using the \textsl{information\_type} attribute and differentiating between the label \textsl{generic\_personal\_information} and other labels specifying the type of data collected. Similarly, this applies to frameworks such as Privacy Shield~\cite{privacyshield} and the GDPR~\cite{eu:gdpr}, where issues such as limiting the data usage purposes should be investigated.

\section{Privacy Icons}
\label{sec:privacy_icons}

\begin{table*}[t]
		\scriptsize
	\centering
	\caption{The list of Disconnect icons with their description, our interpretation, and \framework' queries.}
    \negs
    \begin{tabular}{ @{} L{1.0cm}  L{2.5cm} L{3.3cm} L{4.5cm} L{0.1cm} L{3.8cm}  @{}}  
		\toprule
		\textbf{Icon}  &\textbf{Disconnect Description}  &\textbf{Disconnect Color Assignment} & \textbf{Interpretation as Labels} & & \textbf{Automated Color Assignment} \\
		\midrule
        Expected Use\vspace{0.1cm}\hspace{1cm}
         {\includegraphics[width=0.5\linewidth]{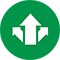}}
  & \scriptsize{Discloses whether data it collects about you is used in ways other than you would reasonably expect given the site's service?} \vspace{-0\baselineskip} & 
\colorbox{red}{\textcolor{white}{\textbf{Red}:}} Yes, w/o choice to opt-out. Or, undisclosed.\newline
\colorbox{yellow}{\textbf{Yellow:}} Yes, with choice to opt-out.\newline
\colorbox{green}{\textbf{Green:}} No.
&  
Let $S$ be the segments with \textbf{category:} \textsl{first-party-collection-use} and \textbf{purpose:} \textsl{advertising}.
&
\multirow{3}{0.2cm}[0pt]{
\[
\begin{drcases}
       \\
       \\
       \\
       \\
       \\
       \\
       \\
       \\
       \\
       \\
       \\           
    \end{drcases}
  \]  
}
&
\multirow{3}{4cm}[-60pt]{ 
\colorbox{yellow}{\textbf{Yellow:}} \underline{All segments in $S$ have}\newline
\textbf{category:} \textsl{user-choice-control} and \textbf{choice-type} $\in$
\newline
 $[$\textsl{opt-in, opt-out-link, opt-out-via-contacting-company}$]$
\colorbox{green}{\textbf{Green:}}  $S= \phi$  \newline
\colorbox{red}{\textcolor{white}{\textbf{Red}:}} Otherwise}
\\ \cmidrule{1-4}
		Expected Collection\vspace{0.1cm}\hspace{1cm}
         {\includegraphics[width=0.5\linewidth]{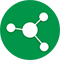}}
  & \scriptsize{Discloses whether it allows other companies like ad providers and analytics firms to track users on the site?} & 
\colorbox{red}{\textcolor{white}{\textbf{Red}:}} Yes, w/o choice to opt-out. Or, undisclosed.\newline
\colorbox{yellow}{\textbf{Yellow:}} Yes, with choice to opt-out.\newline
\colorbox{green}{\textbf{Green:}} No.
&
Let S be the segments with \textbf{category:} \textsl{third-party-sharing-collection}, \textbf{purpose:} $\in[$\textsl{advertising,analytics-research}$]$, and \textbf{action-third-party} $\in [$\textsl{track-on-first-party-website-app,collect-on-first-party-website-app}$]$.
&
&
\\ \cmidrule{1-4}
Precise Location\vspace{0.1cm}\hspace{1cm}
         {\includegraphics[width=0.5\linewidth]{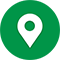}}
  & \scriptsize{Discloses whether the site or service tracks a user's actual geolocation?} & \colorbox{red}{\textcolor{white}{\textbf{Red}:}} Yes, possibly w/o choice.\newline
\colorbox{yellow}{\textbf{Yellow:}} Yes, with choice.\newline
\colorbox{green}{\textbf{Green:}} No.
&	
Let S be the segments with \textbf{personal-information-type:} \textsl{location}.
&   
&

\\	\cmidrule{1-4}
Data Retention\vspace{0.1cm}\hspace{1cm}
         {\includegraphics[width=0.5\linewidth]{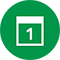}}
  & \scriptsize{ Discloses how long they retain your personal data?} &  
  \colorbox{red}{\textcolor{white}{\textbf{Red}:}} No data retention policy.\newline
\colorbox{yellow}{\textbf{Yellow:}} 12+ months.\newline
\colorbox{green}{\textbf{Green:}} 0-12 months.
&
Let S be the segments with \textbf{category:} \textsl{data-retention}.
&
\multicolumn{2}{L{4.2cm}}
{\colorbox{green}{\textbf{Green:}}\underline{All segments in $S$ have} \textbf{retention-period:} $\in$ \newline
$[$\textsl{stated-period, limited}$]$.\newline
\colorbox{red}{\textcolor{white}{\textbf{Red:}}}  $S= \phi$ \newline \colorbox{yellow}{\textbf{Yellow:}} Otherwise}
\\	 
\cmidrule{1-4}
Children Privacy\vspace{0.1cm}\hspace{1cm}
         {\includegraphics[width=0.5\linewidth]{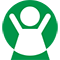}}
  & \scriptsize{Has this website received TrustArc's Children's Privacy Certification?} &  \colorbox{green}{\textbf{Green:}} Yes.
\colorbox{gray}{\textcolor{white}{\textbf{Gray:}}} No.
&
Let S be the segments with \textbf{category:} \textsl{international-and-specific-audiences} and \textbf{audience-type}: \textsl{children}
&
 \multicolumn{2}{L{4.2cm}}
{\colorbox{green}{\textbf{Green:}} $\mbox{length}(S)>0$\newline
\colorbox{red}{\textcolor{white}{\textbf{Red}:}} Otherwise}
\\	
		\bottomrule
	\end{tabular}
	\label{table:disconnect_rules}
\end{table*}

\begin{table}[t]
\scriptsize
\setlength{\tabcolsep}{1pt}
	\caption{Prediction accuracy and $\kappa$ for icon prediction, with the distribution of icons per color based on OPP-115 labels.}
		\centering
		\begin{tabular}{L{1.7cm}C{1.1cm}C{1.0cm}C{1.cm}C{1.0cm}C{0.9cm}C{0.9cm}}
			\toprule
			\textbf{Icon}  & \textbf{Accuracy} & \textbf{Cohen $\kappa$}& \textbf{Hellinger distance} & \colorbox{red}{\textcolor{white}{\textbf{N(R)}}} & \colorbox{green}{\textbf{N(G)} } & \colorbox{yellow}{\textbf{N(Y)}} \\
			\midrule
			Exp. Use& 92\% & 0.76& 0.12 & 41 & 8 & 1\\
			Exp. Collection& 88\% & 0.69& 0.19 & 35 & 12 & 3\\
			Precise Location& 84\% &  0.68 & 0.21& 32 & 14 & 4\\
			Data Retention& 80\%  & 0.63 & 0.13& 29 & 16 & 5\\
			Children Privacy& 98\% & 0.95 & 0.02& 12 & 38 & NA\\
			\bottomrule
            
            \renewcommand{\arraystretch}{1}
		\end{tabular}
	\label{table:icons_accuracy}
\end{table}

Our first application shows the efficacy of \framework in resolving structured queries to privacy policies. 
As a case study, we investigate the Disconnect privacy icons~\cite{disconnect_icons}, described in the first three columns of Table~\ref{table:disconnect_rules}. These icons evolved from a Mozilla-led working group that included the Electronic Frontier Foundation, Center for Democracy and Technology, and the W3C. The database powering these icons originated from TRUSTe (re-branded later as TrustArc), a privacy compliance company, which carried out the task of manually analyzing and labeling privacy policies.

In what follows, we first establish the accuracy of \framework' automatic assignment of privacy icons, using the Disconnect icons as a proof-of-concept. 
We perform a direct comparison between assigning these icons via \framework and assigning them based on annotations by law students~\cite{Wilsonacl16}.
Second, we leverage \framework to investigate the level of permissiveness of the icons that Disconnect assigns based on the TRUSTe dataset. Our findings are consistent with the series of concerns raised around compliance-checking companies over the years~\cite{miyazaki2002internet,caudill2000consumer,pitofsky2000privacy}.
This demonstrates the power of \framework in scalable, automated auditing of privacy compliance checks.

\subsection{Predicting Privacy Icons}
 Given that the rules behind the Disconnect icons are not precisely defined, we translated their description into explicit first-order logic queries to enable automatic processing.  Table~\ref{table:disconnect_rules} shows the original description and color assignment provided by Disconnect. We also show our interpretation of each icon in terms of labels present in the OPP-115 dataset and the automated assignment of colors based on these labels. Our goal is not to reverse-engineer the logic behind the creation of these icons but to show that we can automatically assign such icons with high accuracy, given a plausible interpretation. Hence, this represents our best effort to reproduce the icons, but these rules could easily be adapted as needed. %

To evaluate the efficacy of automatically selecting appropriate privacy icons, we compare the icons produced with \framework' automatic labels to the icons produced based on the law students' annotations from the OPP-115 dataset~\cite{Wilsonacl16}. We perform the evaluation over the same set of 50 privacy policies which we did not use to train \framework (i.e., kept aside as a testing set). Each segment in the OPP-115 dataset has been labeled by three experts. Hence, we take the union of the experts' labels on one hand and the predicted labels from \framework on the other hand. Then, we run the logic presented in Table~\ref{table:disconnect_rules} (Columns 4 and 5) to assign  icons to each policy based on each set of labels.

Table~\ref{table:icons_accuracy} shows the accuracy obtained per icon, measured as the fraction of policies where the icon based on \textit{automatic labels} matched the icon based on the \textit{experts' labels}. The average accuracy across icons is 88.4\%, showing the efficacy of our approach in matching the experts' aggregated annotations. \hmmm{This result is significant in view of Miyazaki and Krishnamurthy's finding~\cite{miyazaki2002internet}: the level of agreement among 3 \textit{trained human judges} assessing privacy policies ranged from 88.3\% to 98.3\%, with an average of 92.7\% agreement overall}. We also show Cohen's $\kappa$, an agreement measure that accounts for agreement due to random chance\footnote{\rm \url{https://en.wikipedia.org/wiki/Cohen\%27s_kappa}}. In our case, the values indicate \textit{substantial} to \textit{almost perfect} agreement~\cite{landis1977measurement}. Finally, we show the distribution of icons based on the \textit{experts' labels} alongside Hellinger distance\footnote{\rm \url{https://en.wikipedia.org/wiki/Hellinger_distance}}, which measures the difference between that distribution and the one produced using the \textit{automatic labels}. This distance assumes small values, illustrating that the distributions are very close. Overall, these results support the potential of automatically assigning privacy icons with \framework.

\subsection{Auditing Compliance Metrics}
Given that we achieve a high accuracy in assigning privacy icons, it is intuitive to investigate how they compare to the icons assigned by Disconnect and TRUSTe. An important consideration in this regard is that several concerns have been raised earlier around the level of leniency of TRUSTe and other compliance companies~\cite{Edelman:2009,trustefb,caudill2000consumer,pitofsky2000privacy}. 
In 2000, the FTC conducted a study on privacy seals, including those of TRUSTe, and found that, of the 27 sites with a privacy seal, approximately only half implemented, at least in part, all four of the fair information practice principles and that only
63\% implemented Notice and Choice. Hence, we pose the following question: \textit{Can we automatically provide evidence of the level of leniency of the Disconnect icons using \framework?} To answer this question, we designed an experiment to compare the icons extracted by \framework' \textit{automatic labels} to the icons assigned by Disconnect on real policies.

One obstacle we faced is that the Disconnect icons have been announced in June 2014~\cite{truste_announce}; many privacy policies have likely been updated since then. To ensure that the privacy policies we consider are within a close time frame to those used by Disconnect, we make use of Ramanath \textit{et al.}'s  ACL/COLING 2014 dataset~\cite{RamanathLSS14}. This dataset contains the body of 1,010 privacy policies extracted between December 2013 and January 2014. We obtained the icons for the same set of sites using the Disconnect privacy icons extension~\cite{disconnect_icons}. Of these, 354 policies had been (at least partially) annotated in the Disconnect dataset. We automatically assign the icons for these sites by passing their policy contents into \framework and applying the rules in Table~\ref{table:disconnect_rules} on the generated \textit{automatic labels}. We report the results for the \textit{Expected Use} and \textit{Expected Collection} icons as they are directly interpretable by \framework. We do not report the rest of the icons because the \textsl{location information} label in the OPP-115 taxonomy included non-precise location (e.g., zip codes), and there was no label that distinguishes the exact retention period. Moreover, the Children privacy icon is assigned through a certification process that does not solely rely on the privacy policy.

\begin{figure*}[t]

\centering

\scriptsize{
}
\begin{minipage}{.33\linewidth}
	\centering
    \begin{subfigure}{.49\linewidth}
	\includegraphics[width=\linewidth]{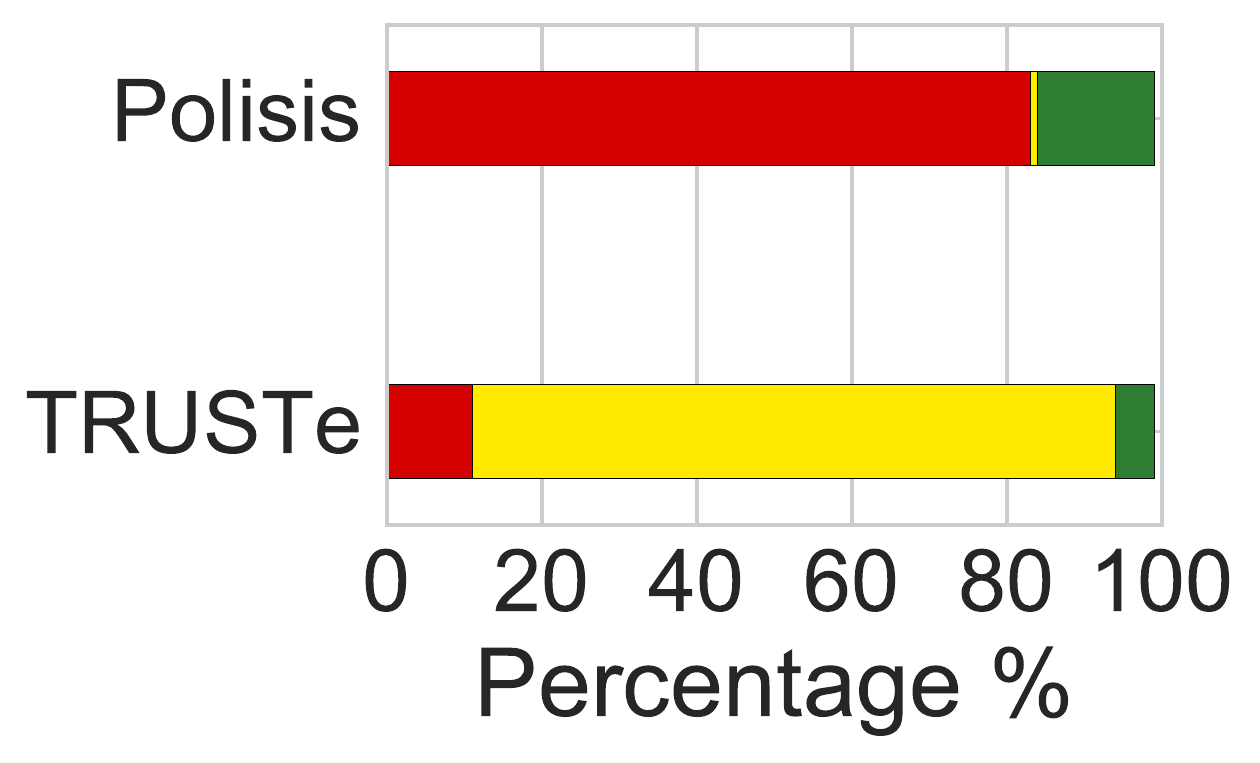}
\caption{Exp. Use}
\label{fig:use_coling_conservative_distribution}
    \end{subfigure}
     \begin{subfigure}{.49\linewidth}
	\includegraphics[width=\linewidth]{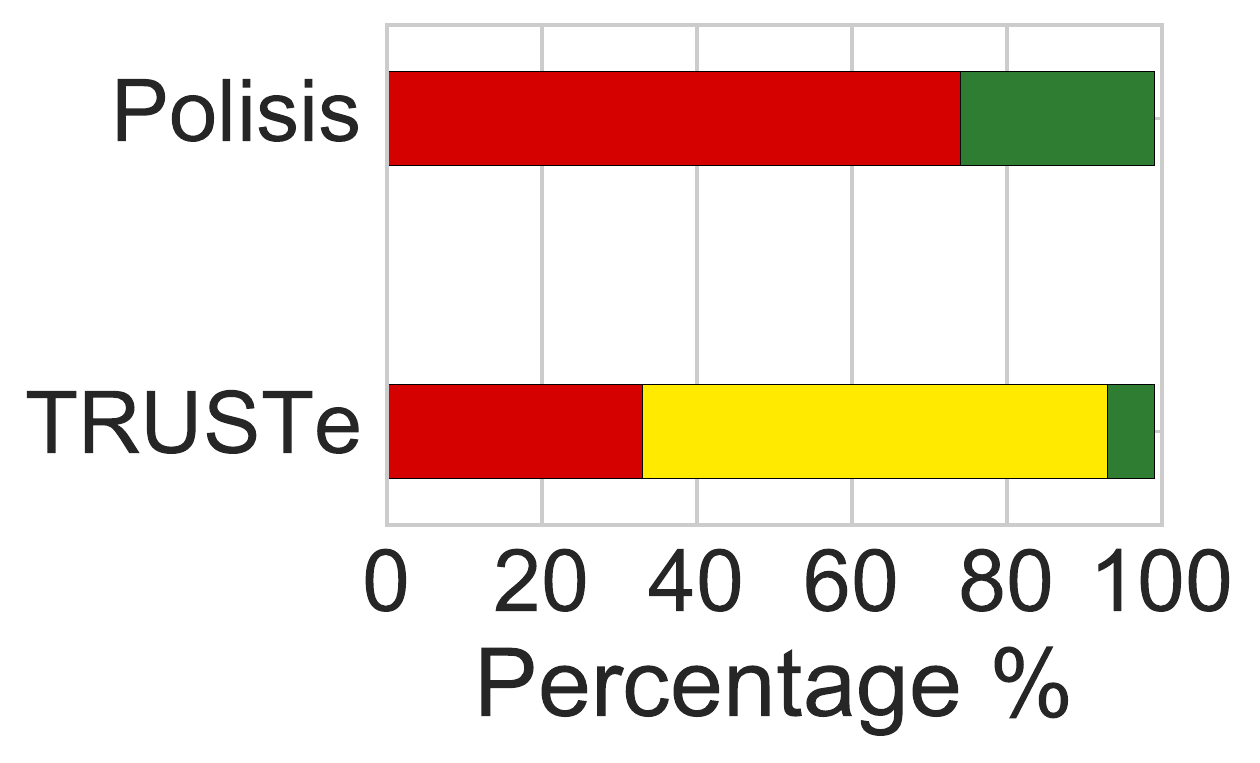}
	\caption{Exp. Collection}
	\label{fig:col_coling_conservative_distribution}
    \end{subfigure}
    \centering
    \caption{Conservative icons' interpretation}
    \label{fig:conservative_coling_icons}
\end{minipage}
\begin{minipage}{.31\linewidth}
	\centering
    \begin{subfigure}{.49\linewidth}
	\includegraphics[width=\linewidth]{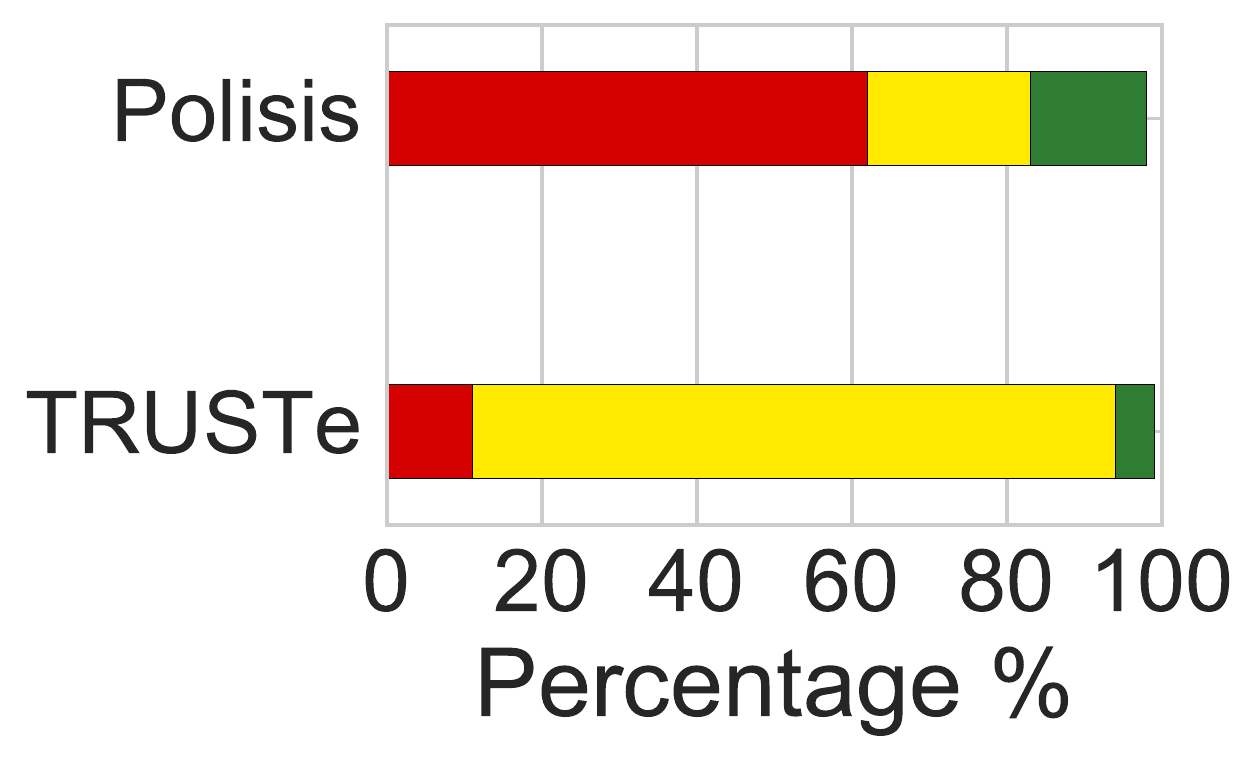}
\caption{Exp. Use}
\label{fig:use_coling_permissive_distribution}
    \end{subfigure}
     \begin{subfigure}{.49\linewidth}
	\includegraphics[width=\linewidth]{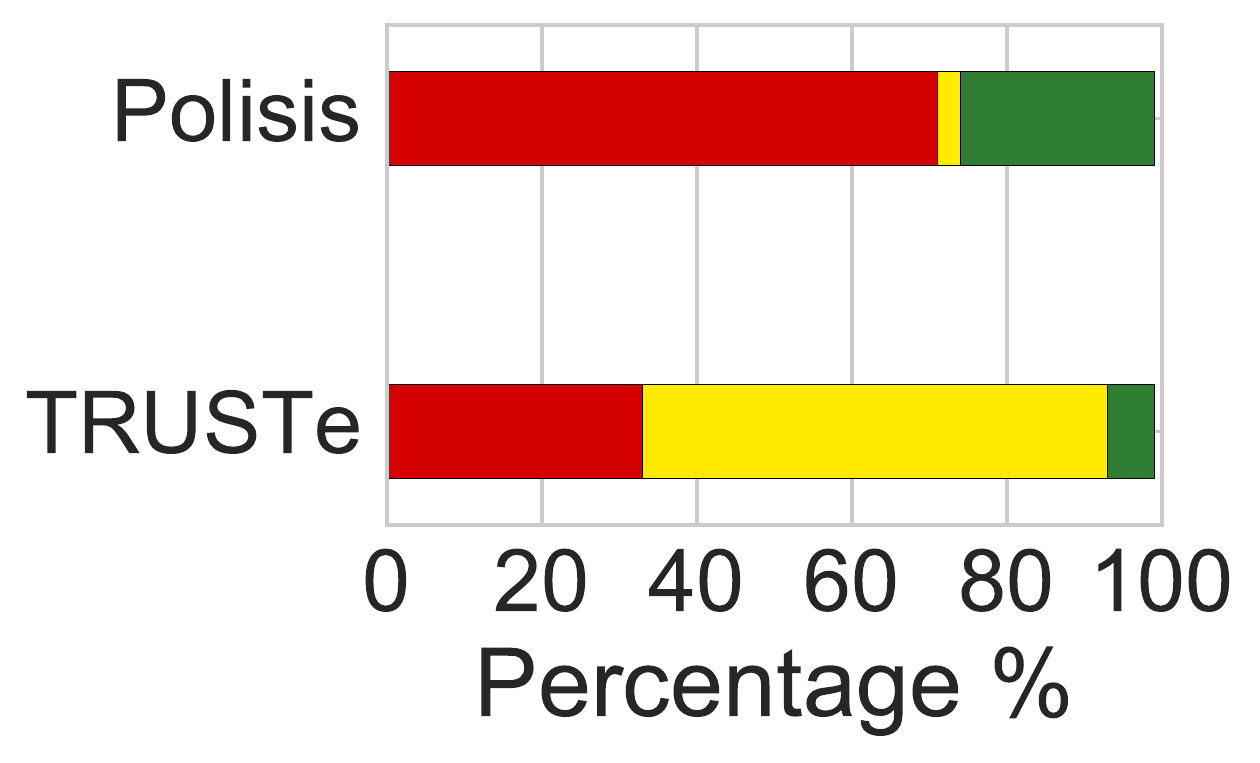}
	\caption{Exp. Collection}
	\label{fig:col_coling_permissive_distribution}
    \end{subfigure}   
     \caption{Permissive icons' interpretation}
    \label{fig:permissive_coling_icons}
\end{minipage}
\begin{minipage}{.35\linewidth}
	\centering
    \begin{subfigure}{.47\linewidth}
	\includegraphics[width=\linewidth]{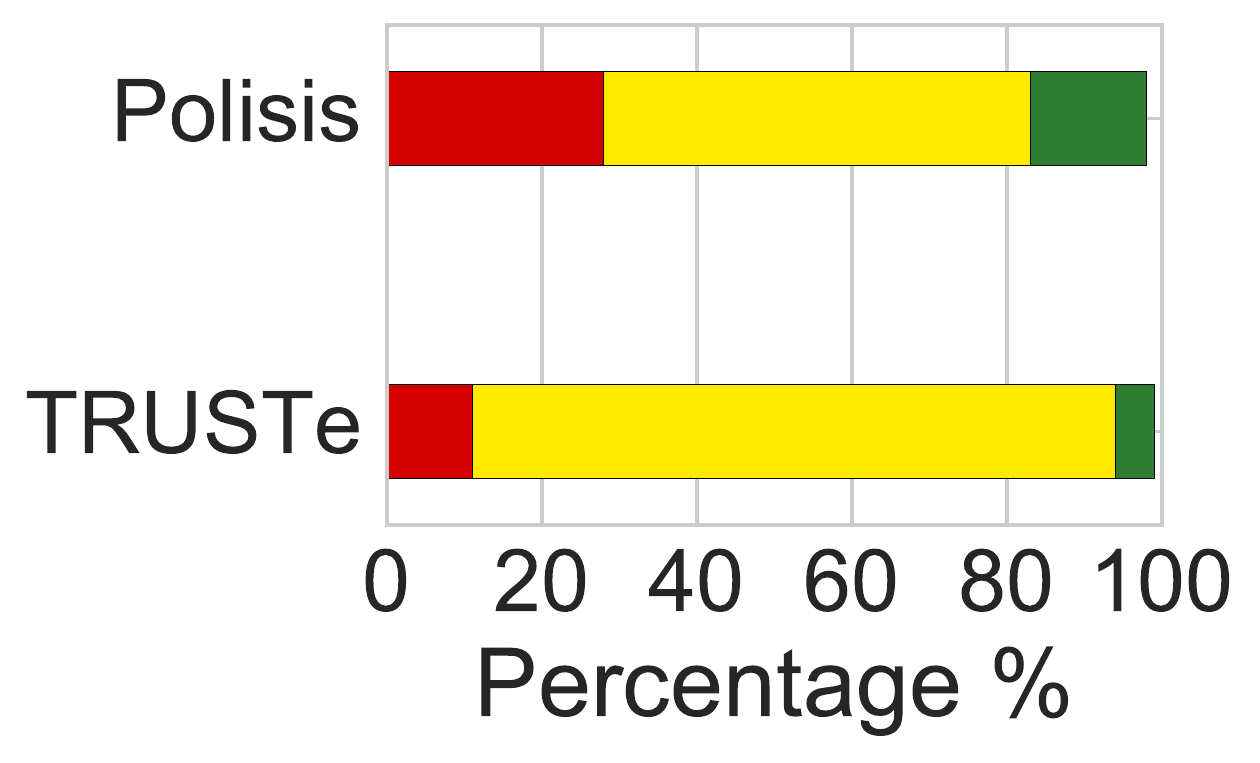}
	\caption{Exp. Use}
	\label{fig:use_coling_very_permissive_distribution}
    \end{subfigure}
     \begin{subfigure}{.47\linewidth}
	\includegraphics[width=\linewidth]{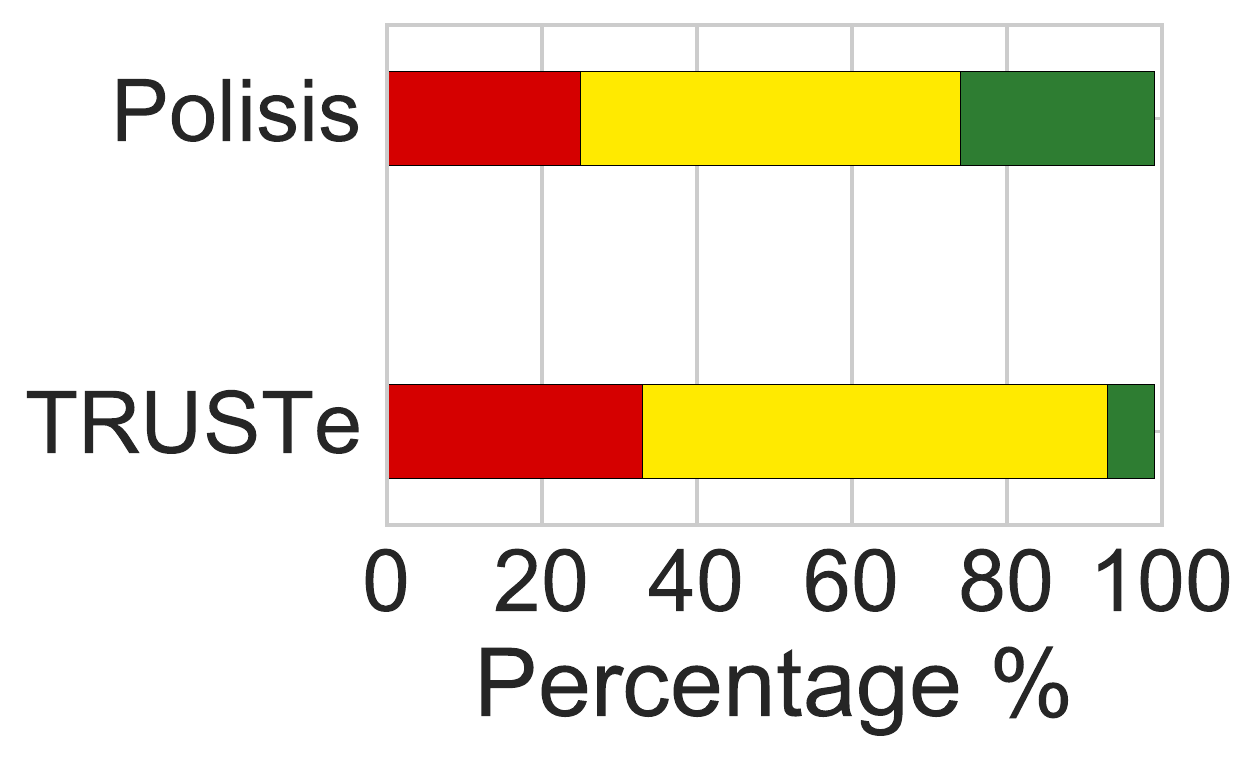}
	\caption{Exp. Collection}
	\label{fig:col_coling_very_permissive_distribution}
    \end{subfigure}
     \caption{Very permissive icons' interpretation}
    \label{fig:very_permissive_coling_icons}
    \end{minipage}
    \vspace{-1\baselineskip}
\end{figure*}

Fig.~\ref{fig:conservative_coling_icons} shows the distribution of automatically extracted icons vs.~the distribution of icons from Disconnect, when they were available. The discrepancy between the two distributions is obvious: the vast majority of the Disconnect icons have a yellow label, indicating that the policies offer the user an opt-out choice (from unexpected use or collection). The Hellinger distances between those distributions are 0.71 and 0.61 for Expected Use and Expected Collection, respectively (i.e., 3--5x the distance in the Table~\ref{table:icons_accuracy}).

This discrepancy might stem from our icon-assignment strategy in Table~\ref{table:disconnect_rules}, where we assign a yellow label only when ``All segments in $S$ (the concerned subset)'' include the opt-in/opt-out choice, which could be considered as conservative. In Fig.~\ref{fig:permissive_coling_icons}, we show the icon distributions when relaxing the yellow-icon condition to become: ``At least one segment in $S$'' includes the opt-in/opt-out choice. \hmmm{Intuitively, this means that the choice segment, when present, should explicitly mention advertising/analytics (depending on the icon type)}. Although the number of yellow icons increases slightly, the icons with the new permissive strategy are significantly red-dominated. The Hellinger distances between those distributions drop to 0.47 and 0.50 for Expected Use and Expected Collection, respectively. This result indicates that the majority of policies do not provide users a choice within the same segments describing data usage for advertising or data collection by third parties.

We go one step further to follow an even more permissive strategy where we assign the yellow label to any policy with $S!=\phi$, given that there is at least one segment in the whole policy (i.e., even outside $S$) with opt-in/opt-out choice. For example, a policy where third-party advertising is mentioned in the middle of the policy while the opt-out choice about another action is mentioned at the end of the policy would still receive a yellow label. The icon distributions, in this case, are illustrated in Fig.~\ref{fig:very_permissive_coling_icons}, with Hellinger distance of 0.22 for Expected Use and 0.19 for Expected Collection. \hmmm{Only in this interpretation of the icons would the distributions of Disconnect and \framework come within reasonable proximity. 
In order to delve more into the factors behind this finding, we conducted a manual analysis of the policies. We found that, due to the way privacy policies are typically written, data collection and sharing are discussed in dedicated parts of the policy, without mentioning user choices. The choices (mostly opt-out) are discussed in a separate section when present, and they cover a small subset of the collected/shared data. In several cases, these choices are neither about the unexpected use (i.e., advertising) nor unexpected collection by third parties (i.e., advertising/analytics). Although our primary hypothesis is that this is due to TRUSTe's database being generally permissive, it can be partially attributed to a potential discrepancy between our versions of analyzed policies and the versions used by TRUSTe (despite our efforts to reduce this discrepancy).}

\subsection{Discussion}

There was no loss of generality when considering only two of the icons; they provided the needed evidence of TRUSTe/TrustArc potentially following a permissive strategy when assigning icons to policies. A developer could still utilize \framework to extract the rest of the icons by either augmenting the existing taxonomy or by performing additional natural language processing on the segments returned by \framework. 
\hmmm{In the vast majority of the cases, whenever the icon definition is to be changed (e.g., to reflect a modification in the regulations), this change can be supported at the rules level, without modifying \framework itself. This is because \framework already predicts a comprehensive set of labels, covering a wide variety of rules.}
Furthermore, by automatically generating icons, we do not intend to push humans completely out of the loop, especially in situations where legal liability issues might arise. \framework can assist human annotators by providing initial answers to their queries and the supporting evidence. In other words, it accurately flags the segments of interest to an annotator's query so that the annotator can make a final decision.

\section{Free-form Question-Answering}
\label{sec:approaches}
Our second application of \framework is \pribot, a system that enables free-form queries (in the form of user questions) on privacy policies. \pribot is primarily motivated by the rise of conversation-first devices, such as voice-activated digital assistants (e.g., Amazon Alexa and Google Assistant) and smartwatches. For these devices, the existing techniques of linking to a privacy policy or reading it aloud are not usable. They might require the user to access privacy-related information and controls on a different device, which is not desirable in the long run~\cite{Schaub:2015}.

To support these new forms of services and the emerging need for automated customer support in this domain~\cite{ibm_customer_support}, we present \pribot as an intuitive and user-friendly method to communicate privacy information. \pribot answers free-form user questions from a previously unseen privacy policy, in real time and with high accuracy. Next, we formalize the problem of free-form privacy \qa and then describe how we leverage \framework to build \pribot.

\subsection{Problem Formulation}
\label{sec:terms}
The input to \pribot consists of a user question $q$ about a privacy policy. \pribot passes $q$ to the ML layer and the policy's link to the Data Layer. The ML layer probabilistically annotates $q$ and each policy's segments with the privacy categories and attribute-value pairs of Fig.~\ref{fig:opp}.

The segments in the privacy policy constitute the pool of candidate answers $\{a_1, a_2,  \ldots, a_M \}$. A subset $\mathcal{G}$ of the answer pool is the ground-truth. We consider an answer $a_k$ as \textit{correct} if $a_k \in \mathcal{G}$ and as \textit{incorrect} if $a_k \notin \mathcal{G}$. If $\mathcal{G}$ is empty, then no answers exist in the privacy policy. 

\subsection{\pribot Ranking Algorithm}
\label{sec:ranking}

\paragraph*{Ranking Score:}
In order to answer the user question, \pribot ranks each potential answer\footnote{For notational simplicity, we henceforth use $a$ to indicate an answer instead of $a_k$.} $a$ by computing a proximity score $s(q,a)$ between $a$ and the question $q$. This is within the \textit{Class Comparison} module of the Application Layer.  To compute $s(q,a)$, we proceed as follows. Given the output of the \textit{Segment Classifier}, an answer is represented as a vector:
\[
\boldsymbol{\alpha}= \{p(c_i|a)^2 \times p(v_j|a) \mid \forall c_i \in \mathcal{C}  , v_j \in \mathcal{V}(c_i)\}
\]
for categories $c_i \in \mathcal{C}$  and values $v_j \in \mathcal{V}(c_i)$ descending from $c_i$. 
Similarly, given the output of the \textit{Query Analyzer}, the question is represented as: 
\[
\boldsymbol{\beta}= \{p(c_i|q)^2 \times p(v_j|q)\mid \forall c_i \in \mathcal{C} , v_j \in \mathcal{V}(c_i)\}
\]
The category probability in both $\boldsymbol{\alpha}$ and $\boldsymbol{\beta}$ is squared to put more weight on the categories at the time of comparison. 
Next, we compute a certainty measure of the answer's high-level categorization. This measure is derived from the entropy of the normalized probability distribution ($p_n$) of the predicted categories:
\begin{equation}
\label{eq:cer_a}
\mbox{$\textit{cer}(a)=1-\left(-\sum\left(p_n(c_i|a) \times \ln(p_n(c_i|a))\right)/\ln(|\mathcal{C}|)\right)$}
\end{equation}

Akin to a dot product between two vectors, we compute the score $s(q,a)$ as:
\begin{equation}
\label{eq:s_q_a}
s(q,a) =\frac{ \sum_i ({\beta}_{i} \times \mbox{min}({\beta}_{i}, {\alpha}_{i}))}{\sum_i {\beta}_{i}^2} \times \textit{cer}(a)
\end{equation}

As answers are typically longer than the question and involve a higher number of significant features, this score prioritizes the answers containing significant features that are also significant in the question. 
The $min$ function and the denominator are used to normalize the score within the range $[0,1]$.

To illustrate the strength of \pribot and its answer-ranking approach, we consider the following question (posed by a Twitter user):\\
{\fontfamily{cmss}\selectfont \small
{``Under what circumstances will you release to 3rd parties?"}
}

Then, we consider two examples of ranked segments by \pribot. The first segment has a ranking score of {0.63}: {\fontfamily{cmss}\selectfont{\small ``Personal information will not be used or disclosed for purposes other than those for which it was collected, except with the consent of the individual or as required by law\ldots"}}\\
The second has a ranking score of {0}: {\fontfamily{cmss}\selectfont\small{``All personal information collected by the TTC will be protected by using appropriate safeguards against loss, theft and unauthorized access, disclosure, copying, use or modification."}}

Although both example segments share terms such as ``personal'' and ``information,'' \pribot ranks them differently. It accounts for the fact that the question and the first segment share the same high-level category: \textsl{$3^{\mbox{rd}}$ Party Collection} while the second segment is categorized under \textsl{Data Security}.

\paragraph*{Confidence Indicator:}

The ranking score is an internal metric that specifies how close each segment is to the question, but does not relay \pribot's certainty in reporting a correct answer to a user. Intuitively, the confidence in an answer should be low when (1) the answer is semantically far from the question (i.e., $s(q,a)$ is low), (2) the question is interpreted ambiguously by \framework, (i.e., classified into multiple high-level categories resulting in a high classification entropy), or (3) when the question contains unknown words (e.g., in a non-English language or with too many spelling mistakes). Taking into consideration these criteria, we compute a confidence indicator as follows:
\begin{equation}
\label{eq:c_q_a}
\textit{conf}(q,a)= s(q,a)*\frac{(\textit{cer}(q)+\mbox{\textit{frac}($q$)})}{2}
\end{equation}
where the categorization certainty measure $\textit{cer}(q)$ is computed similarly to $\textit{cer}(a)$ in Eq.~\eqref{eq:cer_a}, and $s(q,a)$ is computed according to Eq.~\eqref{eq:s_q_a}.
The fraction of known words $\textit{frac}($q$)$ is based on the presence of the question's words in the vocabulary of our \textit{Policies Embeddings'} corpus.

\paragraph*{Potentially Conflicting Answers}

Another challenge is displaying potentially conflicting answers to  users. One answer could describe a general sharing clause while another specifies an exception (e.g., one answer specifies ``share'' and another specifies ``do not share''). To mitigate this issue, we used the same CNN classifier of Sec.~\ref{sec:ml_layer} and exploited the fact that the OPP-115 dataset had optional labels of the form: ``\textsl{does}'' vs. ``\textsl{does not}'' to indicate the presence or absence of sharing/collection. Our classifier had a cross-validation F1 score of 95\%. Hence, we can use this classifier to detect potential discrepancies between the top-ranked answers. The UI of \pribot can thus highlight the potentially conflicting answers to the user.

\section{\pribot Evaluation}
\label{sec:qa_evaluation}
We assess the performance of \pribot  with two metrics: the \textit{predictive accuracy} (Sec.~\ref{sec:acc_evaluation}) of its QA-ranking model and the \textit{user-perceived utility} (Sec.~\ref{sec:user_study}) of the provided answers. This is motivated by research on the evaluation of recommender systems, where the model with the best accuracy is not always rated to be the most helpful by  users%
~\cite{knijnenburg2010receiving}. %

\subsection{Twitter Dataset}
\label{sec:twitter_dataset}

In order to evaluate \pribot with realistic privacy questions, we created a new privacy QA dataset. It is worth noting that we utilize this dataset for the purpose of testing \pribot, not for training it. Our requirements for this dataset were that it (1) must include free-form questions about the privacy policies of different companies and (2) must have a ground-truth answer for each question from the associated policy. 

To this end, we collected, from Twitter, privacy-related questions users had tweeted at companies.
This approach avoids subject bias, which is likely to arise when eliciting privacy-related questions from individuals, who will not pose them out of genuine need. In our collection methodology, we aimed at a QA test set of size between 100 and 200 QA pairs, as is the convention in similar human-annotated QA evaluation domains, such as the Text REtrieval Conference (TREC) and SemEval-2015~\cite{trec:2007,semeval:2015,wang2007jeopardy}.
 
To avoid searching for questions via biased keywords, we started by searching for reply tweets that direct the users to a company's privacy policy (e.g., using queries such as \textsl{"filter:replies our privacy policy"} and \textsl{"filter:replies our privacy statement"}). We then backtracked these reply tweets to the (parent) question tweets asked by customers to obtain a set of 4,743 pairs of tweets, containing privacy questions but also substantial noise due to the backtracking approach. 
Following the best practices of noise reduction in computational social science, we automatically filtered the tweets to keep those containing question marks, at least four words (excluding links, hashtags, mentions, numbers and stop words), and a link to the privacy policy, leaving 260 pairs of question--reply tweets. This is an example of a tweet pair which was removed by the automatic filtering:

\noindent
{\fontfamily{cmss}\selectfont 
\small
\textbf{Question:} ``@Nixxit your site is very suspicious."\\
\textbf{Answer:}		
``@elitelinux Updated it with our privacy policy.  Apologies, but we're not fully up yet and running shoe string."
}

Next, two of the authors independently validated each of the tweets to remove question tweets (a) that were not related to privacy policies, (b) to which the replies are not from the official company account, and (c) with inaccessible privacy policy links in their replies. The level of agreement (Cohen's Kappa) among both annotators for the labels \textsl{valid} vs. \textsl{invalid} was almost perfect ($\kappa=0.84$)~\cite{landis1977measurement}. The two annotators agreed on 231 of the question tweets (of the 260), tagging 182 as \textsl{valid} and 49 as \textsl{invalid}. This is an example of a tweet pair which was annotated as invalid:

\noindent
{\fontfamily{cmss}\selectfont
\small
\textbf{Question:} ``What is your worth then?  You can't do it?  Nuts.''\\
\textbf{Answer:}		
``@skychief26 3/3 You can view our privacy policy at http://t.co/ksmaIK1WaY. Thanks.''
}

This is an example of a tweet pair annotated as valid:\\
{\fontfamily{cmss}\selectfont
\small
\textbf{Question:} ``@myen Are Evernote notes encrypted at rest?''\\
\textbf{Answer:}		
``We're not encrypting at rest, but are encrypting in transit. Check out our Privacy Policy here: http://bit.ly/1tauyfh.''
}

As we wanted to evaluate the answers to these questions with a user study, our estimates of an adequately-sized study led us to randomly sample 120 tweets out of the tweets which both annotators labeled as valid questions. We henceforth refer to them as the \textit{Twitter \qa Dataset}. \hmmm{It is worth mentioning that although our \qa applications extend beyond the Twitter medium, this kind of questions is as close as we can get to testing with the worst-case scenario: informal discourse, with spelling and grammar errors, that is targeted at humans.}

\subsection{\qa Baselines}
\label{sec:qa_baselines}
We compare \pribot's \qa model against three baseline approaches that we developed: (1) \textbf{\retrieval} reflects the state-of-the-art in term-matching retrieval algorithms, (2) \textbf{\etoe} representing a single neural network classifier, and (3) \textbf{\random} as a control approach where questions are answered with random policy segments.

Our first baseline, \retrieval, builds on the BM25 algorithm~\cite{Robertson:2004}, which is the state-of-the-art in ranking models employing term-matching. It has been used successfully across a range of search tasks, such as the TREC evaluations~\cite{beaulieu1997okapi}. We improve on the basic BM25 model by computing the inverse document frequency on the \textit{Policies Corpus} of Sec.~\ref{sec:data_over} instead of a single policy. \retrieval ranks the segments in the policy according to their similarity score with the user's question. This score depends on the presence of distinctive words that link a user's question to an answer.

Our second baseline, \etoe employs a \textit{single} classifier trained to distinguish among all the (mandatory) attribute-values (with $>20$ annotations) from the OPP-115 dataset (81 classes in total). An example segment is {\fontfamily{cmss}\selectfont 
\small
``geographic location information or other location-based information about you and your device''}.
We obtain a micro-average precision of 0.56 (i.e., the classifier is, on average, predicting the right label across the 81 classes in 56\% of the cases -- compared to 3.6\% precision for a random classifier).
After training this model, we extract a ``\textit{semantic vector}'': a representation vector that accounts for the distribution of attribute values in the input text. We extract this vector as the input to the second dense layer (shown Fig.~\ref{fig:class_approach}). \etoe ranks the similarity between a question and a policy segment using the Euclidean distance between semantic vectors. This approach is similar to what has been applied previously in image retrieval, where image representations learned from a large-scale image classification task were effective in visual search applications~\cite{razavian2014visual}.

\subsection{Predictive Accuracy Evaluation}
\label{sec:acc_evaluation}
Here, we evaluate the \textit{predictive accuracy} of \pribot's \qa model by comparing its predicted answers against expert-generated ground-truth answers for the questions of the Twitter \qa Dataset. 

\paragraph{\textbf{Ground-Truth Generation:}} 

Two of the authors generated the ground-truth answers to the questions from the Twitter \qa Dataset. They were given a user's question (tweet) and the segments of the corresponding policy. Each policy consists of 45 segments on average (\textit{min=12, max=344, std=37}). 
Each annotator selected \textit{independently}, the subset of these segments which they consider as best responding to the user's question. 
This annotation took place \textit{prior} to generating the answers using our models to avoid any bias. While deciding on the answers, the annotators accounted for the fact that multiple segments of the policy might answer a question. 

After finishing the individual annotations, the two annotators consolidated the differences in their labels to reach an agreed-on set of segments; each assumed to be answering the question. We call this the \textit{ground-truth} set for each question. 
The annotators agreed on at least one answer in 88\% of the questions for which they found matching segments, thus signifying a substantial overlap. Cohen's $\kappa$, measuring the agreement on one or more answer, was $0.65$, indicating substantial agreement~\cite{landis1977measurement}. 
We release this dataset, comprising the questions, the policy segments, and the ground-truth answers per question \mbox{at \href{https://pribot.org/data.html}{\url{https://pribot.org/data.html}}}.

\begin{figure}[t]%
\centering
\scriptsize{
\fcolorbox{black}{lightpink}{\rule{0pt}{4pt}\rule{2pt}{0pt}} \random \quad \fcolorbox{black}{orchid} {\rule{0pt}{4pt}\rule{2pt}{0pt}} \retrieval \fcolorbox{black}{darkorchid} {\rule{0pt}{4pt}\rule{2pt}{0pt}} \etoe \fcolorbox{black}{midnightblue} {\rule{0pt}{4pt}\rule{2pt}{0pt}} \pribot}

	\begin{subfigure}{0.49\columnwidth}
		\centering
		\includegraphics[width=\columnwidth]{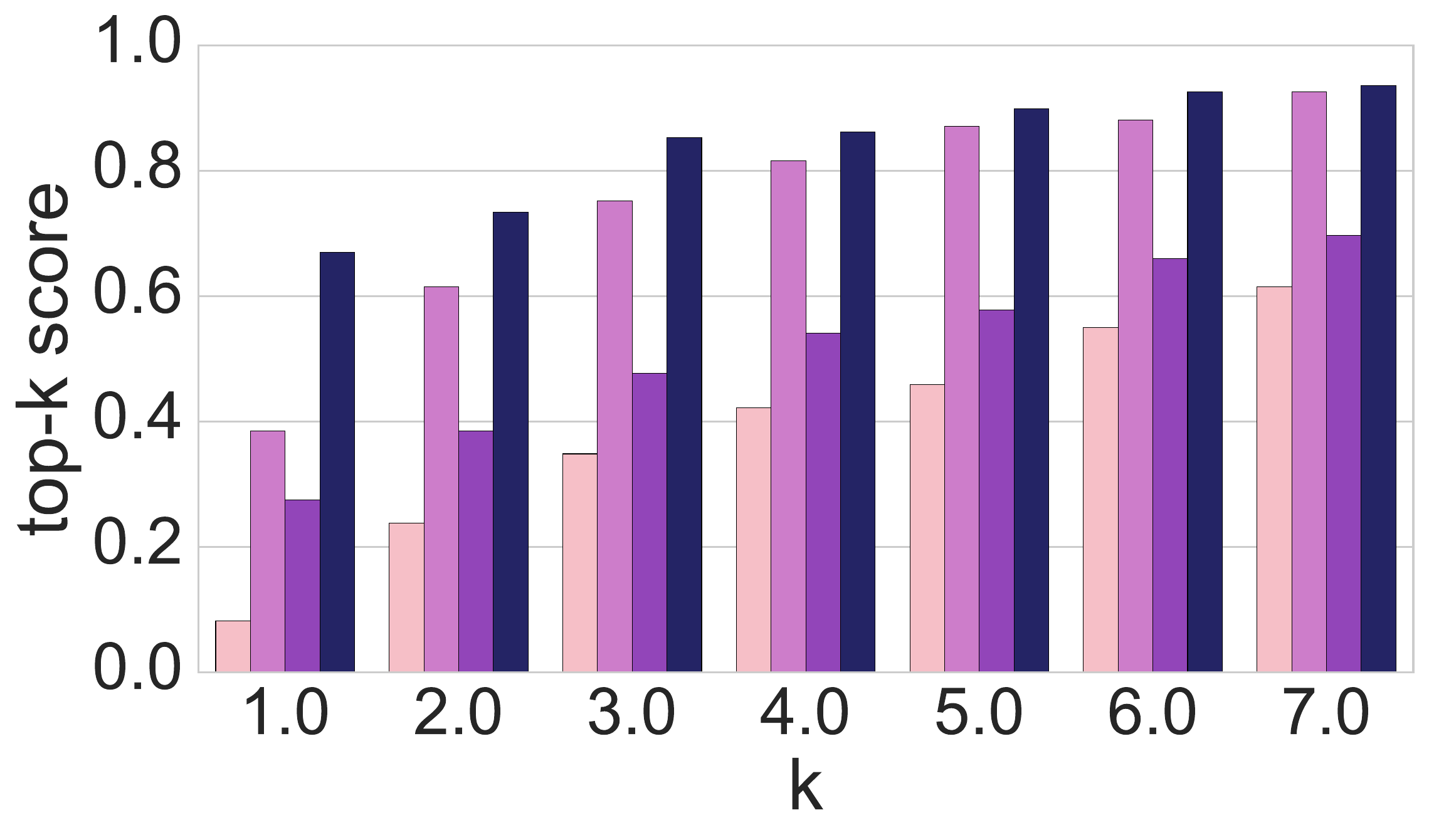}%
        \negs
		\caption{\topk}%
		\label{fig:top_k_score_reconciled}%
	\end{subfigure}
	\begin{subfigure}{0.49\columnwidth}
		\centering
	\includegraphics[width=\columnwidth]{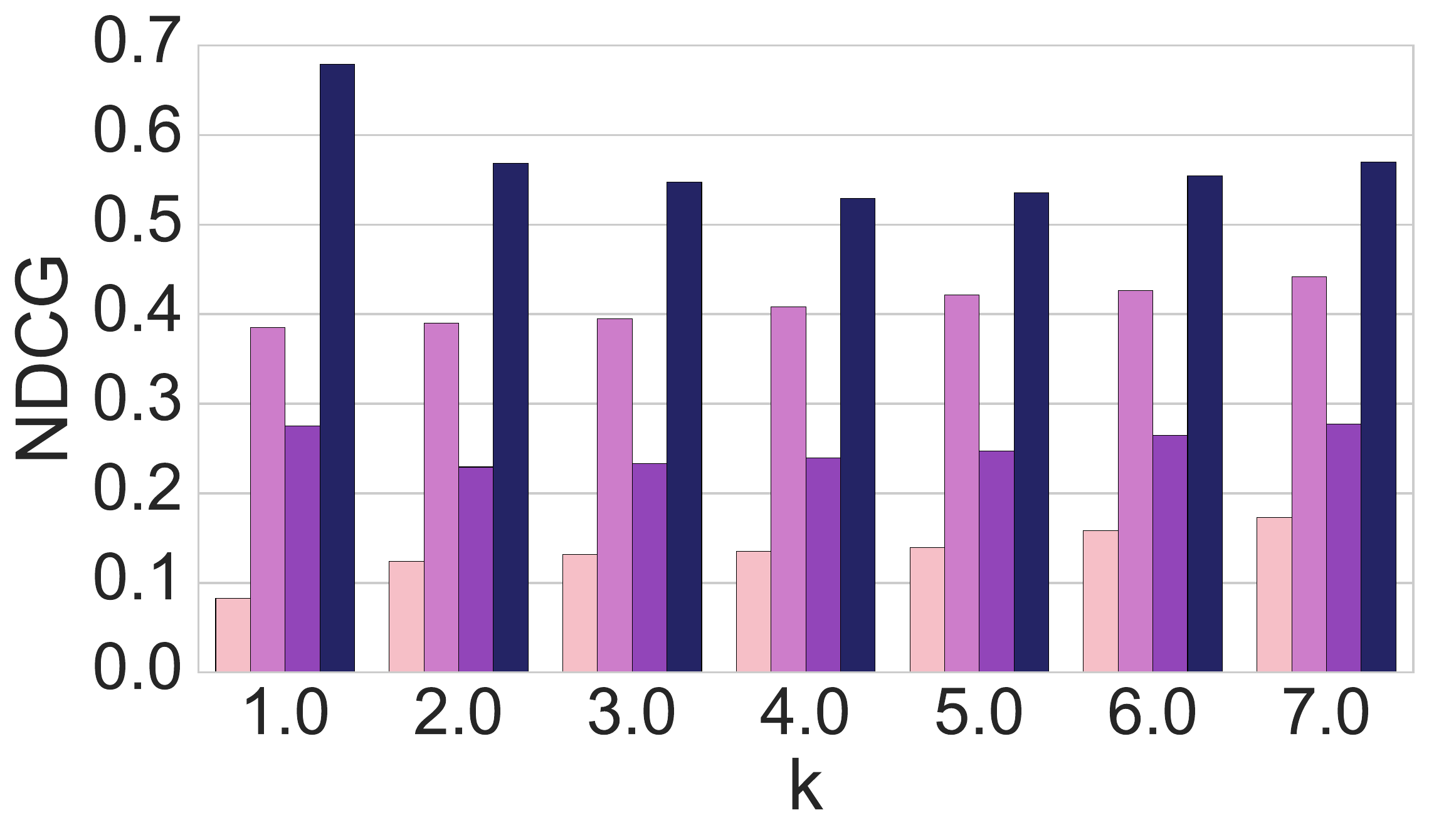}%
    \negs
		\caption{\NDCG}%
		\label{fig:ndcg_reconciled}%
	\end{subfigure}
       \caption{Accuracy metrics as a function of $k$.}
\end{figure}

We then generated, for each question, the predicted ranked list of answers according to each \qa model (\pribot and the other three baselines). In what follows, we evaluate the predictive accuracy of these models.

\paragraph{\textbf{Top-$k$ Score:}}
\label{sec:qametrics}

We first report the \topk, a widely used and easily interpretable metric, which denotes the portion of questions having at least one correct answer in the top $k$ returned answers. It is desirable to achieve a high \topk for low values of $k$ so that the user has to process less information before reaching a correct answer.  Fig.~\ref{fig:top_k_score_reconciled} shows how the \topk varies as a function of $k$. \pribot's model has the best performance over the other three models by a large margin, especially at the low values of $k$. 
For example, at $k=1$, \pribot has a \topk of 0.68, which is significantly larger than the scores of 0.39 (\retrieval), 0.27 (\etoe), and 0.08 (\random) (\mbox{$p\mbox{-value}<0.05$} according to pairwise Fisher's exact test, corrected with Bonferroni method for multiple comparisons).
\pribot further reaches a \topk of 0.75, 0.82, and 0.87 for $k\in\{2,3,4\}$. To put these numbers in the wider context of free-form \qa systems, we note that the top-1 accuracy reported by IBM Watson's team on a large insurance domain dataset (a training set of 12,889 questions and 21,325 answers) was 0.65 in 2015~\cite{FengXGWZ15} and was later improved to 0.69 in 2016~\cite{tan2016improved}. Given that \pribot had to overcome the absence of publicly available \qa datasets, our top-1 accuracy value of 0.68 is on par with such systems.
We also observe that the \retrieval model outperforms the \etoe model. This result is not entirely surprising since we seeded \retrieval with a large corpus of 130K unsupervised policies, thus improving its performance on answers with matching terms. 

\paragraph{\textbf{Policy Length}}
We now assess the impact of the policy length on \pribot's accuracy.
\text{First}, we report the \textit{Normalized Discounted Cumulative Gain (NDCG)}~\cite{jarvelin2002cumulated}. Intuitively, it indicates that a relevant document's usefulness decreases logarithmically with the rank. This metric captures how presenting the users with more choices affects their user experience as they need to process more text. Also, it is not biased by the length of the policy. The $DCG$ part of the metric is computed as ${{DCG_{k}} =\sum _{i=1}^{k}{\frac {rel_{i}}{\log _{2}(i+1)}}}$, where $rel_{i}$ is 1 if answer $a_i$ is correct and 0 otherwise. NDCG at $k$ is obtained by normalizing the $DCG_{k}$ with the maximum possible $DCG_{k}$ across all values of $k$. We show in Fig.~\ref{fig:ndcg_reconciled} the average NDCG across questions for each value of $k$. It is clear that \pribot's model consistently exhibits superior NDCG.
This indicates that \pribot is poised to perform better in a system where low values of $k$ matter the most.

\begin{figure}[t]%
	
	\begin{minipage}{0.48\columnwidth}
    \centering
    \tiny{
\fcolorbox{black}{lightpink}{\rule{0pt}{4pt}\rule{0.3pt}{0pt}} \random
\fcolorbox{black}{orchid} {\rule{0pt}{4pt}\rule{0.3pt}{0pt}} \retrieval 
\fcolorbox{black}{darkorchid} {\rule{0pt}{4pt}\rule{0.3pt}{0pt}} \etoe 
\fcolorbox{black}{midnightblue} {\rule{0pt}{4pt}\rule{0.3pt}{0pt}} \pribot} 
\includegraphics[width=\textwidth]{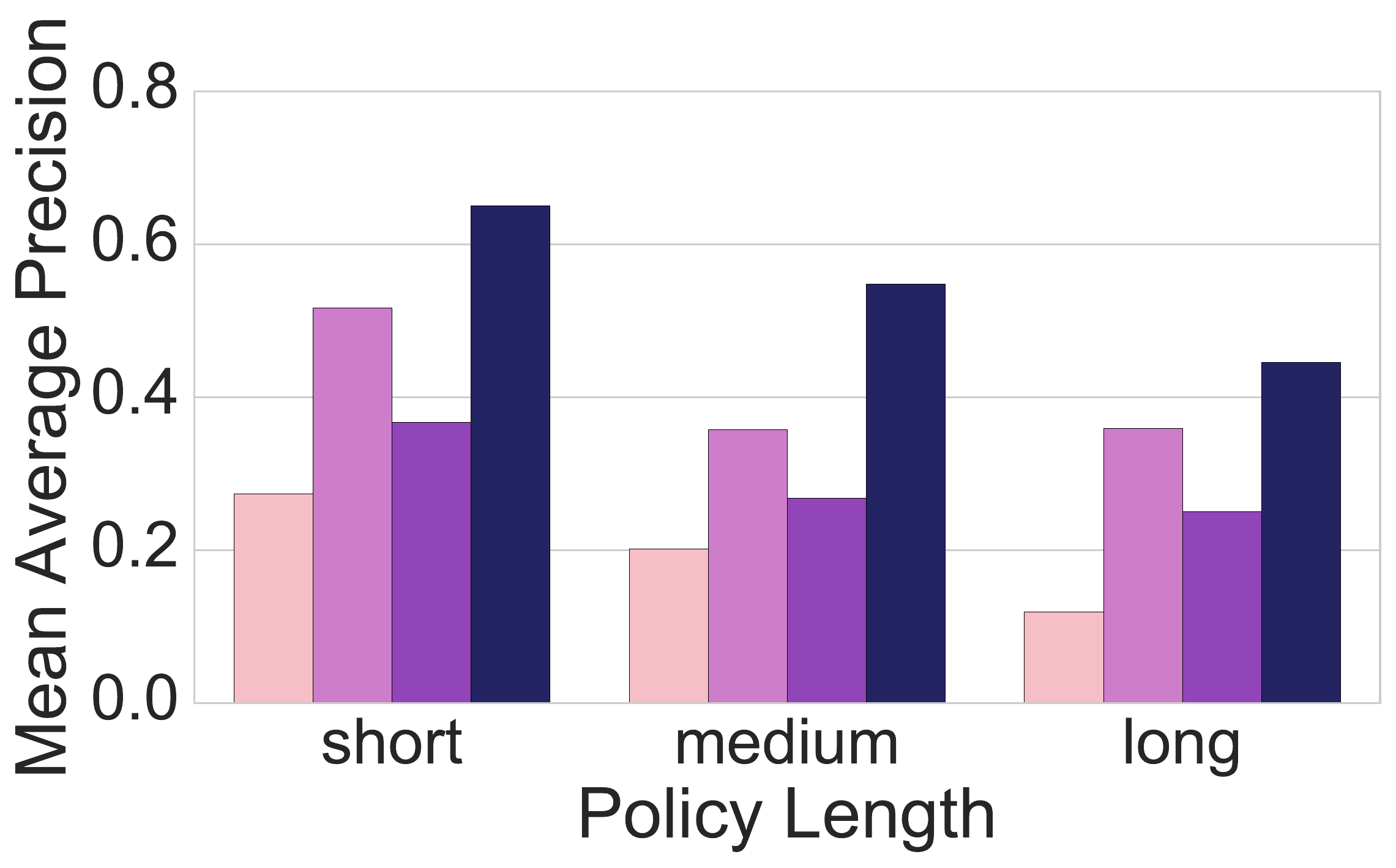}%
\caption{Variation of \MAP across policy lengths.}%
\label{fig:map_evolution__level_1}%
	\end{minipage}
    \begin{minipage}{0.47\columnwidth}
    \centering
\tiny{
\fcolorbox{black}{green1}{\rule{0pt}{4pt}\rule{0.3pt}{0pt}} \fontfamily{cmss}\selectfont{WP-NoSub}  \fcolorbox{black}{green2} {\rule{0pt}{4pt}\rule{0.3pt}{0pt}} \wpEmb \fcolorbox{black}{green3} {\rule{0pt}{4pt}\rule{0.3pt}{0pt}} \peNoSubEmb  \fcolorbox{black}{green4} {\rule{0pt}{4pt}\rule{0.3pt}{0pt}} \peEmb}
		\centering
		\includegraphics[width=\textwidth]{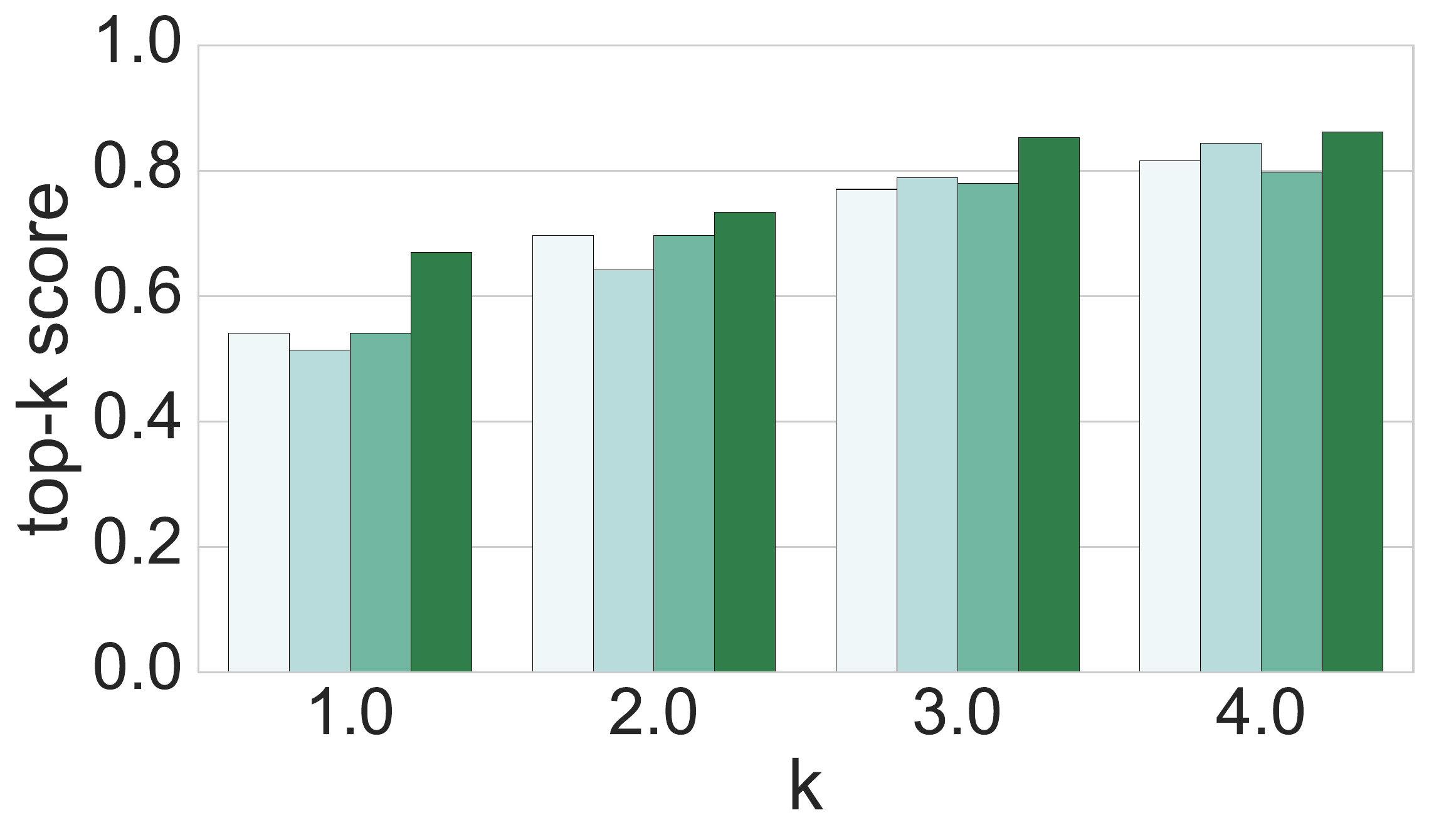}%
        \negs
		\caption{\topk of \pribot with different pre-trained embeddings.}%
	\label{fig:embeddings_top_k_score_reconciled_q1_}%
	\end{minipage}
    \negs
\end{figure}

Second, to further focus on the effect of policy length, we categorize the policy lengths ($\#{\mbox{\textit{segments}}}$) into \textit{short}, \textit{medium}, and \textit{high}, based on the 33rd and the 66th percentiles (i.e., corresponding to $\#{\mbox{\textit{segments}}}$ of 28 and 46). We then compute a metric independent of $k$, namely, the Mean Average Precision (\MAP), which is the mean of the area under the precision-recall curve across all questions. Informally, \MAP is an indicator of whether all the correct answers get ranked highly. We see from Fig.~\ref{fig:map_evolution__level_1} that, for short policies, the \retrieval model is within 15\% of the \MAP of \pribot's model, which makes sense given the smaller number of potential answers. With medium-sized policies, \pribot's model is better by a large margin. This margin is still considerable with long policies.

\paragraph{\textbf{Confidence Indicator}} 
Comparing the confidence (using the indicator from Eq.~\eqref{eq:c_q_a}) of incorrect answers predicted by \pribot (mean=0.37, variance=0.04) with the confidence of correct answers (mean=0.49, variance =0.05) shows that \pribot places lower confidence in the answers that turn out to be incorrect. Hence, we can use the confidence indicator to filter out the incorrect answers. For example, by setting the condition: $\textit{conf}(q,a) \geq 0.6$ to accept \pribot's answers, we can enhance the top-1 accuracy to 70\%. 
This indicator delivers another advantage: its components are independently interpretable by the application logic. If the score $s(q,a)$ of the top-1 answer is too low, the user can be notified that the policy might not contain an answer to the question. A low value of $\textit{cer}(q)$ indicates that the user might have asked an ambiguous question; the system can ask the user back for a clarification. 

\paragraph{\textbf{Pre-trained Embeddings Choice}}
As discussed in Sec.~\ref{sec:ml_layer}, we utilize our custom Policies Embeddings, which have the two properties of (1) being domain-specific and (2) using subword embeddings to handle out-of-vocabulary words. We test the efficacy of this choice by studying three variants of pre-trained embeddings.
For the first variant, we start from our Policies Embeddings ({\peEmb}), and we disable the subwords mode, thus only satisfying the first property; we call it \peNoSubEmb. The second variant is the \textit{fastText} Wikipedia Embeddings from~\cite{fasttext_pretrained}, trained on the English Wikipedia, thus only satisfying the second property; we denote it as {\wpEmb}. The third variant is {\wpEmb}, with the subword mode disabled, thus satisfying neither property; we call it {\wpNoSubEmb}. In Fig.~\ref{fig:embeddings_top_k_score_reconciled_q1_}, we show the \topk of \pribot on our Twitter \qa dataset with each of the four pre-trained embeddings.  First, we can see that our Policies Embeddings outperform the other models for all values of $k$, scoring 14\% and 5\% more than the closest variant at $k=1$ and $k=2$, respectively. 
As expected, the domain-specific model without subwords embeddings (\peNoSubEmb) has a weaker performance by a significant margin, especially for the top-1 answer. 
Interestingly, the difference is much narrower between the two Wikipedia embeddings since their vocabulary already covers more than 2.5M tokens.
Hence, subword embeddings play a less pronounced role there. In sum, the advantage of using subwords embeddings with the \peEmb model originates from their domain specificity and their ability to compensate for the missing words from the vocabulary. 

\subsection{User-Perceived Utility Evaluation}
\label{sec:user_study}
We conducted a user study to assess the \textit{user-perceived utility} of the automatically generated answers. This assessment was done for each of the four different conditions (\retrieval, \etoe, \pribot and \random). We evaluated the top-3 responses of each \qa approach to each question. Thus, we assess the utility of 360 answers to 120 questions per approach.

\paragraph{Study Design:}
We used a between-subject design by constructing four surveys, each corresponding to a different evaluation condition.  We display a series of 17 QA pairs (each on a different page). Of these, 15 are a random subset of the pool of 360 QA pairs (of the evaluated condition) such that a participant does not receive two \qa pairs with the same question. The other two questions are randomly positioned anchor questions serving as attention checkers. Additionally, we enforce a minimum duration of 15 seconds for the respondent to evaluate each \qa pair, with no maximum duration enforced. We include an open-ended Cloze reading comprehension test~\cite{cefr}; we used the test to  weed out the responses with a low score, indicating a poor reading skill. 

\paragraph{\textbf{Participant Recruitment:}}
After obtaining an IRB approval, we recruited 700 Amazon MTurk workers with previous success rate \textit{$>$95\%}, to complete our survey. With this number of users, each \qa pair received evaluations from at least 7 different individuals. We compensated each respondent with \$2.
With an average completion time of 14 minutes, this makes the average pay around \$8.6 per hour (US Federal minimum wage is \$7.25). While not fully representative of the general population, our set of participants exhibited high intra-group diversity, but little difference across the respondent groups. Across all respondents, the average age is 34 years (\textit{std=10.5}), 62\% are males, 38\% are females, more than 82\% are from North America, more than 87\% have some level of college education, and more than 88\% reported being employed.

\begin{figure}
\centering
  \includegraphics[width=0.9\linewidth]{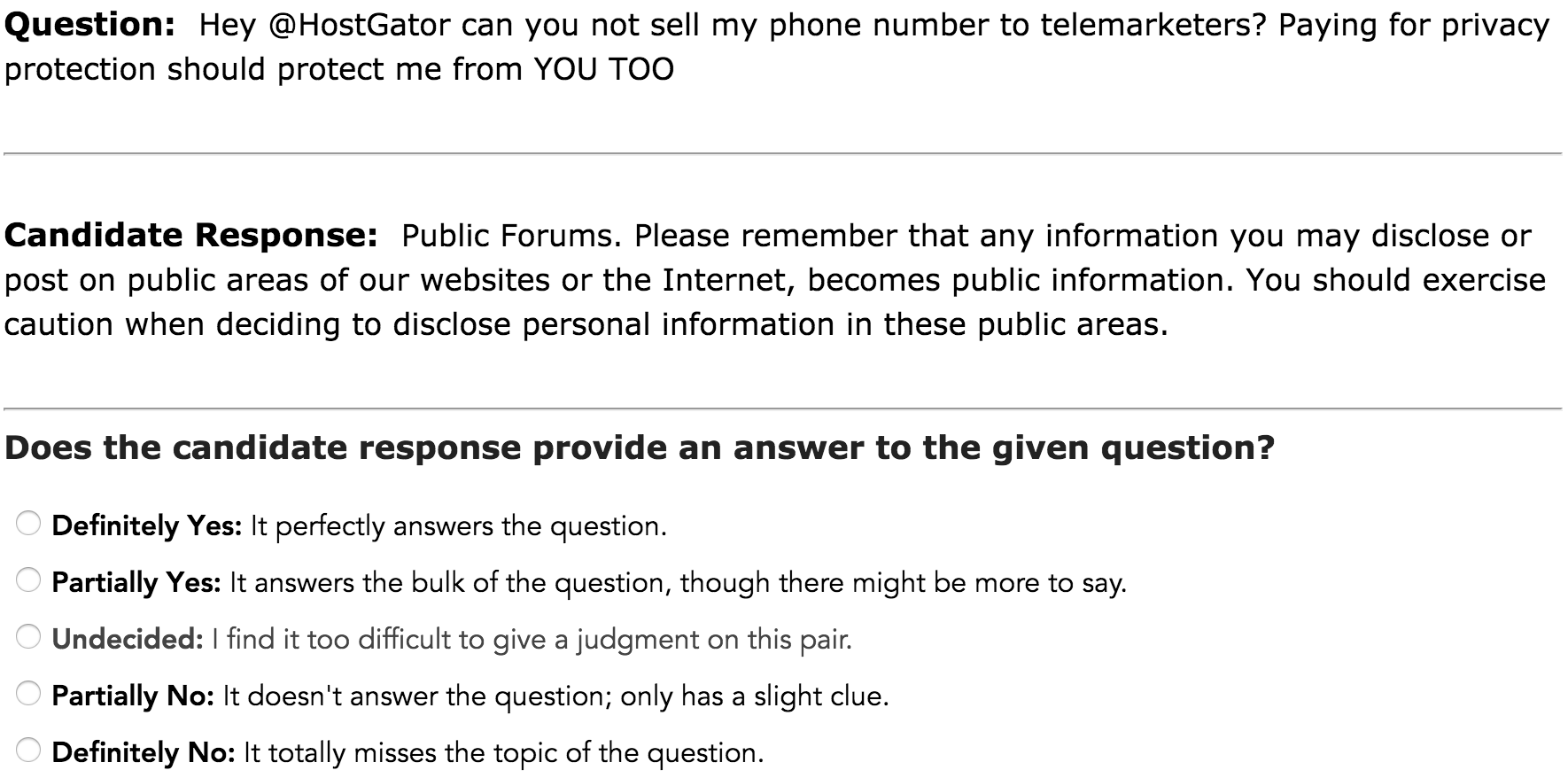}
  \caption{An example of a \qa pair displayed to the respondents.}
  \label{fig:qa_example}
\end{figure}

\paragraph{\qa Pair Evaluation:}
To evaluate the relevance for a \qa pair, we display the question and the candidate answer as shown in Fig.~\ref{fig:qa_example}.
We asked the respondents to rate whether the candidate response provides an answer to the question on a 5-point Likert scale (1=\textit{Definitely Yes} to 5=\textit{Definitely No}), as evident in Fig.~\ref{fig:qa_example}. We denote a respondent's evaluation of a \textbf{single} candidate answer corresponding to a \qa pair as relevant (irrelevant) if s/he chooses either \textit{Definitely Yes (Definitely No)} or \textit{Partially Yes (Partially No)}. \kmmm{We consolidate the evaluations of multiple users per answer by following the methodology outlined in similar studies~\cite{Wilson:2016}, which consider the answer as relevant if labeled as relevant by a certain fraction of users. We took this fraction as 50\% to ensure a majority agreement. Generally, we observed the respondents to agree on the relevance of the answers. Highly mixed responses, where 45--55\% of the workers tagged the answer as relevant, constituted less than 16\% of the cases.}

\begin{table}[t]
\renewcommand{\arraystretch}{1.05}
\scriptsize
\centering
\vspace{\baselineskip}
\caption{top-$k$ relevance score by evaluation group.}
\vspace{-0.5\baselineskip}
\begin{tabular}{lcccc}
\multirow{3}{*}{\textbf{Group}} & \multirow{3}{*}{\textbf{\textit{N}}} & \multicolumn{3}{c}{\textbf{\texttt{top-$k$ Relevance Score}}}\\
\cmidrule{3-5}
& & $k=1$ & $k=2$ & $k=3$ \\
\midrule
\random  & 180 & 0.37 & 0.59 & 0.76 \\
\retrieval & 184 & 0.46 & 0.71 & 0.79\\
\etoe  & 153 & 0.48 & 0.71 & 0.85 \\
\pribot & 183 & \textbf{0.70} & \textbf{0.78} & \textbf{0.89}\\

\bottomrule
\end{tabular}
\label{table:rel_eval}
\end{table}

\paragraph{User Study Results:}
As in the previous section, we compute the \topk for relevance (i.e., the portion of questions having at least one user-relevant answer in the top $k$ returned answers). Table~\ref{table:rel_eval} shows this score for the four \qa approaches with $k \in \{1,2,3\}$, where \pribot clearly outperforms the three baseline approaches. The respondents regarded at least one of the top-3 answers as relevant for 89\% of the questions, with the first answer being relevant in 70\% of the cases. In comparison, for $k=1$, the scores were 46\% and 48\% for the \retrieval and the \etoe models respectively (\mbox{$p\mbox{-value}<=0.05$} according to pairwise Fisher’s exact test, corrected with Holm-Bonferroni method for multiple comparisons).
An avid reader might notice some differences between the predictive models' accuracy (Section~\ref{sec:acc_evaluation}) and the users' perceived quality. This is actually consistent with the observations from research in recommender systems where the prediction accuracy does not always match user's satisfaction~\cite{knijnenburg2010receiving}. For example, the \topk metric for accuracy differs by 2\%, -3\%, and 6\% with respect to the perceived relevance in the \pribot model.
Another example is that the \etoe model and the \retrieval have smaller differences in this study than Sec.~\ref{sec:qametrics}. We conjecture that the score shift with \etoe model is due to some users accepting answers which match the question's topic even when the actual details of the answer are irrelevant.

\section{Discussion}
\label{sec:discussion}
\paragraph*{Limitations}
\framework might be limited by the employed privacy taxonomy. Although the OPP-115 taxonomy covers a wide variety of privacy practices~\cite{Wilsonacl16}, there are certain types of applications that it does not fully capture. One mitigation is to use \framework as an initial step in order to filter the relevant data at a high level before applying additional, application-specific text processing. Another mitigation is to leverage \framework' modularity by amending it with new categories/attributes and training these new classes on the relevant annotated dataset.

Moreover, \framework, like any automated approach, exhibits instances of misclassification that should be accounted for in any application building on it. One way to mitigate this problem is using confidence scores, similar to that of Eq.~\eqref{eq:c_q_a} to convey the (un)certainty of a reported result, whether it is an answer, an icon, or another form of short notice.
Last but not least, \framework is not guaranteed to be robust in handling an adversarially constructed privacy policy. An adversary could include valid and meaningful statements in the privacy policy, carefully crafted to mislead \framework' automated classifiers. For example, an adversary can replace words, in the policy, with synonyms that are far in our embeddings space. While the modified policy has the same meaning, \framework might misclassify the modified segments.

\paragraph*{Deployment:}
We provide three prototype web applications for end-users. The first is an application that visualizes the different aspects in the privacy policy, powered by the annotations from \framework (available as a web application and a browser extension for Chrome and Firefox).
The second is a chatbot implementation of \pribot for answering questions about privacy policies in a conversational interface. The third is an application for extracting the privacy labels from several policies, given their links. These applications are available at \siteurl. 

\paragraph*{Legal Aspects}
We also want to stress the fact that \framework \textit{is not intended }to replace the legally-binding privacy policy. Rather, it offers a complementary interface for privacy stakeholders to easily inquire the contents of a privacy policy. Following the trend of automation in legal advice~\cite{legalbots}, insurance claim resolution~\cite{insurancebots}, and privacy policy presentation~\cite{liu2016modeling,zimmeck2014privee}, third parties, such as automated legal services firms or regulators, can deploy \framework as a solution for their users. As is the standard in such situations, these parties should amend \framework with a disclaimer specifying that it is based on automatic analysis and does not represent the actual service provider~\cite{lawsrobots}. 

\new{Companies and service providers can internally deploy an application similar to \pribot as an assistance tool for their customer support agents to handle privacy-related inquiries. Putting the human in the loop allows for a favorable trade-off between the utility of \framework and its legal implications. 
For a wider discussion on the issues surrounding automated legal analysis, we refer the interested reader to the works of McGinnis and Pearce~\cite{mcginnis2014legal} and Pasquale~\cite{pasquale2015four}.}

\paragraph{Privacy-Specificity of the Approach: } Finally, our approach is uniquely tailored to the privacy domain both from the data perspective and from the model-hierarchy perspective. However, we envision that applications with similar needs would benefit from extensions of our approach, both on the classification level and the QA level.

\section{Related Work}
\label{sec:related}

\paragraph{Privacy Policy Analysis:}
There have been numerous attempts to create easy-to-navigate and alternative presentations of privacy policies. Kelley \textit{et al.}~\cite{kelley2009nutrition} studied using nutrition labels as a paradigm for displaying privacy notices. Icons representing the privacy policies have also been proposed~\cite{cranor2006user,holtz2011privacy}. Others have proposed standards to push service providers to encode privacy policies in a machine-readable format, such as P3P~\cite{cranor2002web}, but they have not been adopted by browser developers and service providers. \framework has the potential to automate the generation of a lot of these notices, without relying on the respective parties to do it themselves.

Recently, several researchers have explored the potential of automated analysis of privacy policies. For example, Liu \textit{et al.}~\cite{liu2016modeling} have used deep learning to model the vagueness of words in privacy policies. 
Zimmeck \textit{et al.}~\cite{zimmeck2016automated} have been able to show significant inconsistencies between app practices and their privacy policies via automated analysis. These studies, among others~\cite{liu2016analyzing,sathyendra2016automatic}, have been largely enabled by the release of the OPP-115 dataset by Wilson \textit{et al.}~\cite{Wilsonacl16}, containing 115 privacy policies extensively annotated by law students. 
Our work is the first to provide a generic system for the automated analysis of privacy policies. In terms of the comprehensiveness and the accuracy of the approach, \framework makes a major improvement over the state of the art. It allows transitioning from labeling of policies with a few practices (e.g., the works by Zimmeck and Bellovin~\cite{zimmeck2014privee} and Sathyendra \textit{et al.}~\cite{sathyendra2017identifying}) to a much more fine-grained annotation (up to 10 high-level and 122 fine-grained classes), thus enabling a richer set of applications.

\paragraph{Evaluating the Compliance Industry:}
Regulators and researchers are continuously scrutinizing the practices of the privacy compliance industry~\cite{miyazaki2002internet,caudill2000consumer,pitofsky2000privacy}. Miyazaki and Krishnamurthy~\cite{miyazaki2002internet} found no support that participating in a seal program is an indicator of following privacy practice standards. The FTC has found discrepancies between the practical behaviors of the companies, as reported in their privacy policies, and the privacy seals they have been granted~\cite{pitofsky2000privacy}. \framework can be used by these researchers and regulators to automatically, and continuously perform such checks at scale. It can provide the initial evidence that could be processed by skilled experts afterward, thus reducing the analysis time and the cost.

\paragraph{Automated Question Answering:}
Our \qa system, \pribot, is focused on \textit{non-factoid} questions, which are usually complex and open-ended. Over the past few years, deep learning has yielded superior results to traditional retrieval techniques in this domain~\cite{FengXGWZ15,tan2016improved,Rao:2016}. Our main contribution is that we build a \qa system, without a dataset that includes questions and answers, while achieving results on par with the state of the art on other domains. We envision that our approach could be transplanted to other problems that face similar issues.

\section{Conclusion}
\label{sec:conclusion}
We proposed \framework, the first generic framework that enables detailed automatic analysis of privacy policies. It can assist users, researchers, and regulators in processing and understanding the content of privacy policies at scale. To build \framework, we developed a new hierarchy of neural networks that extracts both high-level privacy practices as well as fine-grained information from privacy policies. Using this extracted information, \framework enables several applications. In this paper, we demonstrated two applications: structured and free-form querying. In the first example, we use \framework' output to extract short notices from the privacy policy in the form of privacy icons and to audit TRUSTe's policy analysis approach. In the second example, we build \pribot that answers users' free-form questions in real time and with high accuracy. Our evaluation of both applications reveals that \framework matches the accuracy of expert analysis of privacy policies. Besides these applications, \framework opens opportunities for further innovative privacy policy presentation mechanisms, including summarizing policies into simpler language. It can also enable comparative shopping applications that advise the consumer by comparing the privacy aspects of multiple applications they want to choose from.

\section*{Acknowledgements}
\normalsize {This research was partially funded by the Wisconsin Alumni Research Foundation and the US National Science Foundation under grant agreements CNS-1330596 and CNS-1646130. }
\Urlmuskip=0mu plus 1mu
{\footnotesize \bibliographystyle{IEEEtran}
\bibliography{z_references}}

\appendix

\section*{Appendix A: Full Classification Results}
\label{sec:appendix_full_results}
\setlength{\textfloatsep}{1pt plus 1.0pt minus 1.0pt}
\setlength{\intextsep}{1pt plus 1.0pt minus 1.0pt}
We present the classification results at the category level for the Segment Classifier and at 15 selected attribute levels, using the hyperparameters of Table ~\ref{table:class_cat}. 
\vspace{0.3\baselineskip}

\begin{minipage}[t]{\linewidth}
	\scriptsize
	\centering
	\begin{tabular*}{8cm}{lccC{0.5cm} C{0.7cm} C{0.8cm}}
        \toprule
		\multicolumn{6}{c}{Classification results at the category level for the Segment Classifier}\\
		\midrule
		\textbf{Label}  &\textbf{Prec.}  &\textbf{Recall} & \textbf{F1}& \textbf{Top-1 Prec.}& \textbf{Support}\\
		\midrule
 Data Retention & 0.83 & 0.66 & 0.71 & 0.68 & 88\\
Data Security & 0.88 & 0.83 & 0.85 & 0.79 & 201\\
Do Not Track  & 0.94 & 0.97 & 0.95 & 0.88 & 16\\
$1^{\mbox{st}}$ Party Collection & 0.79 & 0.79 & 0.79 & 0.79 & 1211\\
Specific Audiences & 0.96 & 0.94 & 0.95 & 0.93 & 156\\
Introductory/Generic & 0.81 & 0.66 & 0.70 & 0.75 & 369\\
Policy Change & 0.95 & 0.84 & 0.88 & 0.93 & 112\\
Non-covered Practice & 0.76 & 0.67 & 0.70 & 0.60 & 280\\
Privacy Contact Info & 0.90 & 0.85 & 0.87 & 0.88 & 137\\
$3^{\mbox{rd}}$ Party Sharing & 0.79 & 0.80 & 0.79 & 0.82 & 908\\
Access, Edit, Delete & 0.89 & 0.75 & 0.80 & 0.87 & 133\\
User Choice/Control & 0.74 & 0.74 & 0.74 & 0.69 & 433\\
\midrule
Average & 0.85 & 0.79 & 0.81 & 0.80\\
\bottomrule
	\end{tabular*}
	\AppendixCaption{Classification results at the category level for the Segment Classifier}
	\label{table:}
	\scriptsize
	\centering
	\vspace{1\baselineskip}\AppendixCaption{Classification results for attribute: \textsl{change-type}}
	\begin{tabular}{L{3.3cm}ccC{0.5cm} C{0.9cm} C{0.9cm}}
        	\multicolumn{5}{c}{Classification results for attribute: \textsl{change-type}}\\ 		\midrule
		\textbf{Label}  &\textbf{Prec.}  &\textbf{Recall} & \textbf{F1}&  \textbf{Support}\\
		\midrule
 privacy-relevant-change & 0.78 & 0.76 & 0.77 &  77\\
unspecified & 0.79 & 0.76 & 0.76 & 90\\
\midrule
Average & 0.78 & 0.76 & 0.76 \\
\bottomrule
	\end{tabular}
	\label{table:change-type}
	\scriptsize
	\centering
	\vspace{1\baselineskip}\AppendixCaption{Classification results for attribute: \textsl{notification-type}}
	\begin{tabular}{L{3.3cm}ccC{0.5cm} C{0.9cm} C{0.9cm}}
        	\multicolumn{5}{c}{Classification results for attribute: \textsl{notification-type}}\\ 		\midrule
		\textbf{Label}  &\textbf{Prec.}  &\textbf{Recall} & \textbf{F1}&  \textbf{Support}\\
		\midrule
 general-notice-in-privacy-policy & 0.80 & 0.77 & 0.78 & 76\\
general-notice-on-website & 0.64 & 0.62 & 0.62  & 52\\
personal-notice & 0.69 & 0.66 & 0.67  & 50\\
unspecified & 0.81 & 0.72 & 0.75  & 24\\
\midrule
Average & 0.73 & 0.69 & 0.71 \\
\bottomrule
	\end{tabular}
	\label{table:notification-type}

\end{minipage}

\begin{minipage}[t]{\linewidth}

	\scriptsize
	\centering
	\vspace{1\baselineskip}\AppendixCaption{Classification results for attribute: \textsl{do-not-track-policy}}
	\begin{tabular}{L{3.3cm}ccC{0.5cm} C{0.9cm} C{0.9cm}}
        	\multicolumn{5}{c}{Classification results for attribute: \textsl{do-not-track-policy}}\\ 		\midrule
		\textbf{Label}  &\textbf{Prec.}  &\textbf{Recall} & \textbf{F1}&  \textbf{Support}\\
		\midrule
 honored & 1.00 & 1.00 & 1.00 & 8\\
not-honored & 1.00 & 1.00 & 1.00  & 26\\
\midrule
Average & 1.00 & 1.00 & 1.00 \\
\bottomrule
	\end{tabular}
	\label{table:do-not-track-policy}
	\scriptsize
	\centering
	\vspace{1\baselineskip}\AppendixCaption{Classification results for attribute: \textsl{security-measure}}
	\begin{tabular}{L{3.3cm}ccC{0.5cm} C{0.9cm} C{0.9cm}}
        	\multicolumn{5}{c}{Classification results for attribute: \textsl{security-measure}}\\ 		
        	\midrule
		\textbf{Label}  &\textbf{Prec.}  &\textbf{Recall} & \textbf{F1}& \textbf{Support}\\
		\midrule
 data-access-limitation & 0.89 & 0.78 & 0.81 &  35\\
generic & 0.84 & 0.83 & 0.83 & 102\\
privacy-review-audit & 0.97 & 0.58 & 0.62 & 13\\
privacy-security-program & 0.87 & 0.69 & 0.73 & 31\\
secure-data-storage & 0.82 & 0.64 & 0.69 &  17\\
secure-data-transfer & 0.91 & 0.80 & 0.84 & 26\\
secure-user-authentication & 0.97 & 0.58 & 0.63 & 12\\
\midrule
Average & 0.90 & 0.70 & 0.74  \\
\bottomrule
	\end{tabular}
	\label{table:security-measure}
	\scriptsize
	\centering
	\begin{tabular}{L{3.3cm}ccC{0.5cm} C{0.9cm} C{0.9cm}}
        	\multicolumn{5}{c}{Classification results for attribute: \textsl{personal-information-type}}\\ 		
        	\midrule
		\textbf{Label}  &\textbf{Prec.}  &\textbf{Recall} & \textbf{F1}&  \textbf{Support}\\
		\midrule
 computer-information & 0.84 & 0.80 & 0.82  & 88\\
contact & 0.90 & 0.89 & 0.90 &  342\\
cookies-and-tracking-elements & 0.95 & 0.92 & 0.94  & 272\\
demographic & 0.93 & 0.90 & 0.92 & 86\\
financial & 0.89 & 0.86 & 0.87 & 99\\
generic-personal-information & 0.82 & 0.79 & 0.80  & 441\\
health & 1.00 & 0.56 & 0.61 &  8\\
ip-address-and-device-ids & 0.93 & 0.93 & 0.93 & 104\\
location & 0.88 & 0.88 & 0.88 & 107\\
personal-identifier & 0.67 & 0.61 & 0.63 &  31\\
social-media-data & 0.73 & 0.84 & 0.78 & 23\\
survey-data & 0.77 & 0.86 & 0.81 & 22\\
unspecified & 0.71 & 0.70 & 0.71 & 456\\
user-online-activities & 0.80 & 0.82 & 0.81 & 224\\
user-profile & 0.79 & 0.68 & 0.72  & 96\\
\midrule
Average & 0.84 & 0.80 & 0.81 \\
\bottomrule
	\end{tabular}
	\vspace{1\baselineskip}\AppendixCaption{Classification results for attribute: \textsl{personal-information-type}}
	\label{table:personal-information-type}
	\scriptsize
	\centering
	\vspace{1\baselineskip}\AppendixCaption{Classification results for attribute: \textsl{purpose}}
	\begin{tabular}{L{3.3cm}ccC{0.5cm} C{0.9cm} C{0.9cm}}
        	\multicolumn{5}{c}{Classification results for attribute: \textsl{purpose}}\\ 		\midrule
		\textbf{Label}  &\textbf{Prec.}  &\textbf{Recall} & \textbf{F1}&  \textbf{Support}\\
		\midrule
 additional-service-feature & 0.75 & 0.76 & 0.75 & 374\\
advertising & 0.92 & 0.91 & 0.92  & 286\\
analytics-research & 0.88 & 0.86 & 0.87  & 239\\
basic-service-feature & 0.76 & 0.73 & 0.74 & 401\\
legal-requirement & 0.92 & 0.91 & 0.91 &  79\\
marketing & 0.86 & 0.83 & 0.84 &  312\\
merger-acquisition & 0.95 & 0.96 & 0.95 & 38\\
personalization-customization & 0.79 & 0.80 & 0.80 &  149\\
service-operation-and-security & 0.81 & 0.77 & 0.79 & 200\\
unspecified & 0.72 & 0.68 & 0.70 & 249\\
\midrule
Average & 0.84 & 0.82 & 0.83 \\
\bottomrule
	\end{tabular}
	\label{table:purpose}
	\scriptsize
	\centering
	\vspace{1\baselineskip}\AppendixCaption{Classification results for attribute: \textsl{choice-type}}
	\begin{tabular}{L{3.3cm}ccC{0.5cm} C{0.9cm} C{0.9cm}}
        	\multicolumn{5}{c}{Classification results for attribute: \textsl{choice-type}}\\ 		\midrule
		\textbf{Label}  &\textbf{Prec.}  &\textbf{Recall} & \textbf{F1}&  \textbf{Support}\\
		\midrule
 browser-device-privacy-controls & 0.89 & 0.95 & 0.92  & 171\\
dont-use-service-feature & 0.69 & 0.65 & 0.67 & 213\\
first-party-privacy-controls & 0.75 & 0.62 & 0.66  & 71\\
opt-in & 0.78 & 0.81 & 0.79 & 406\\
opt-out-link & 0.82 & 0.74 & 0.77 &  167\\
opt-out-via-contacting-company & 0.87 & 0.81 & 0.84 & 127\\
third-party-privacy-controls & 0.82 & 0.62 & 0.66 & 99\\
unspecified & 0.65 & 0.54 & 0.56 & 117\\
\midrule
Average & 0.78 & 0.72 & 0.73 \\
\bottomrule
	\end{tabular}
	\label{table:choice-type}

\end{minipage}

\begin{minipage}[t]{\linewidth}

	\scriptsize
	\centering
	\vspace{1\baselineskip}\AppendixCaption{Classification results for attribute: \textsl{action-third-party}}
	\begin{tabular}{L{3.3cm}ccC{0.5cm} C{0.9cm} C{0.9cm}}
        	\multicolumn{5}{c}{Classification results for attribute: \textsl{third-party-entity}}\\ 		\midrule
		\textbf{Label}  &\textbf{Prec.}  &\textbf{Recall} & \textbf{F1}& \textbf{Support}\\
		\midrule
collect-on-first-party-website-app & 0.78 & 0.64 & 0.68 & 113\\
receive-shared-with & 0.87 & 0.87 & 0.87 & 843\\
see & 0.83 & 0.79 & 0.81 & 63\\
track-on-first-party-website-app & 0.75 & 0.86 & 0.79  & 107\\
unspecified & 0.60 & 0.51 & 0.52  & 57\\
\midrule
Average & 0.77 & 0.74 & 0.73 \\
\bottomrule
	\end{tabular}
	\label{table:action-third-party}
	\scriptsize
	\centering
	\vspace{1\baselineskip}\AppendixCaption{Classification results for attribute: \textsl{access-type}}
	\begin{tabular}{L{3.3cm}ccC{0.5cm} C{0.9cm} C{0.9cm}}
        	\multicolumn{5}{c}{Classification results for attribute: \textsl{access-type}}\\ 		\midrule
		\textbf{Label}  &\textbf{Prec.}  &\textbf{Recall} & \textbf{F1}&  \textbf{Support}\\
		\midrule
 edit-information & 0.65 & 0.62 & 0.63 & 172\\
unspecified & 0.98 & 0.64 & 0.71 & 14\\
view & 0.55 & 0.53 & 0.53 & 47\\
\midrule
Average & 0.73 & 0.60 & 0.62 \\
\bottomrule
	\end{tabular}
	\label{table:access-type}
	\scriptsize
	\centering
	\vspace{1\baselineskip}\AppendixCaption{Classification results for attribute: \textsl{audience-type}}
	\begin{tabular}{L{3.3cm}ccC{0.5cm} C{0.9cm} C{0.9cm}}
        	\multicolumn{5}{c}{Classification results for attribute: \textsl{audience-type}}\\ 		\midrule
		\textbf{Label}  &\textbf{Prec.}  &\textbf{Recall} & \textbf{F1}&  \textbf{Support}\\
		\midrule
 californians & 0.98 & 0.97 & 0.98& 60\\
children & 0.98 & 0.97 & 0.97 & 161\\
europeans & 0.97 & 0.95 & 0.96 & 23\\
\midrule
Average & 0.98 & 0.97 & 0.97 \\
\bottomrule
	\end{tabular}
	\label{table:audience-type}
	\scriptsize
	\centering
	\vspace{1\baselineskip}\AppendixCaption{Classification results for attribute: \textsl{choice-scope}}
	\begin{tabular}{L{3.3cm}ccC{0.5cm} C{0.9cm} C{0.9cm}}
        	\multicolumn{5}{c}{Classification results for attribute: \textsl{choice-scope}}\\ 		\midrule
		\textbf{Label}  &\textbf{Prec.}  &\textbf{Recall} & \textbf{F1}&  \textbf{Support}\\
		\midrule
 both & 0.61 & 0.53 & 0.54 & 71\\
collection & 0.74 & 0.68 & 0.70  & 302\\
first-party-collection & 0.63 & 0.55 & 0.56  & 109\\
first-party-use & 0.80 & 0.68 & 0.71 &  236\\
third-party-sharing-collection & 0.81 & 0.60 & 0.64  & 98\\
third-party-use & 0.57 & 0.51 & 0.50 &  60\\
unspecified & 0.55 & 0.55 & 0.55 &  76\\
use & 0.62 & 0.55 & 0.56 &  140\\
\midrule
Average & 0.67 & 0.58 & 0.59 \\
\bottomrule
	\end{tabular}
	\label{table:choice-scope}
	\renewcommand{\arraystretch}{1.05}
	\scriptsize
	\centering
	\vspace{1\baselineskip}\AppendixCaption{Classification results for attribute: \textsl{action-first-party}}
	\begin{tabular}{L{3.3cm}ccC{0.5cm} C{0.9cm} C{0.9cm}}
        	\multicolumn{5}{c}{Classification results for attribute: \textsl{action-first-party}}\\ 		\midrule
		\textbf{Label}  &\textbf{Prec.}  &\textbf{Recall} & \textbf{F1}& \textbf{Support}\\
		\midrule
 collect-in-mobile-app & 0.84 & 0.75 & 0.79 & 68\\
collect-on-mobile-website & 0.58 & 0.54 & 0.56 & 11\\
collect-on-website & 0.65 & 0.65 & 0.65 & 739\\
unspecified & 0.61 & 0.60 & 0.60 & 294\\
\midrule
Average & 0.67 & 0.64 & 0.65 \\
\bottomrule
	\end{tabular}
	\label{table:action-first-party}
	\scriptsize
	\centering
	\vspace{1\baselineskip}\AppendixCaption{Classification results for attribute: \textsl{does-does-not}}
	\begin{tabular}{L{3.3cm} ccC{0.5cm} C{0.9cm} C{0.9cm}}
		\multicolumn{5}{c}{Classification results for attribute: \textsl{does-does-not}}\\
		\midrule
		\textbf{Label}  &\textbf{Prec.}  &\textbf{Recall} & \textbf{F1}& \textbf{Support}\\
		\midrule
 does & 0.82 & 0.93 & 0.86& 1436\\
does-not & 0.82 & 0.93 & 0.86  & 200\\
\midrule
Average & 0.82 & 0.93 & 0.86\\
\bottomrule
	\end{tabular}
	\label{table:}
	\scriptsize
	\centering
	\vspace{1\baselineskip}\AppendixCaption{Classification results for attribute: \textsl{retention-period}}
	\begin{tabular}{L{3.3cm}ccC{0.5cm} C{0.9cm} C{0.9cm}}
        	\multicolumn{5}{c}{Classification results for attribute: \textsl{retention-period}}\\ 		\midrule
		\textbf{Label}  &\textbf{Prec.}  &\textbf{Recall} & \textbf{F1}&  \textbf{Support}\\
		\midrule
 indefinitely & 0.45 & 0.48 & 0.47 & 8\\
limited & 0.74 & 0.75 & 0.75 & 27\\
stated-period & 0.94 & 0.94 & 0.94  & 10\\
unspecified & 0.82 & 0.77 & 0.77  & 41\\
\midrule
Average & 0.74 & 0.74 & 0.73 \\
\bottomrule
	\end{tabular}
	\label{table:retention-period}

\end{minipage}

\begin{minipage}[t]{\linewidth}

\scriptsize
\centering
\vspace{1\baselineskip}\AppendixCaption{Classification results for attribute: \textsl{identifiability}}
\begin{tabular}{L{3.3cm}ccC{0.5cm} C{0.9cm} C{0.9cm}}
	\multicolumn{5}{c}{Classification results for attribute: \textsl{identifiability}}\\ 		\midrule
	\textbf{Label}  &\textbf{Prec.}  &\textbf{Recall} & \textbf{F1}&  \textbf{Support}\\
	\midrule
	aggregated-or-anonymized & 0.89 & 0.89 & 0.89 & 284\\
	identifiable & 0.81 & 0.81 & 0.81 & 492\\
	unspecified & 0.63 & 0.63 & 0.63 & 98\\
	\midrule
	Average & 0.77 & 0.78 & 0.77\\
	\bottomrule
\end{tabular}
\label{table:identifiability}

\end{minipage}

\iftrue
\vspace{3 \baselineskip}
\section*{Appendix B: Applications' Screenshots}
\label{sec:appendix-chatbot}
In this appendix, we first show screenshots of \pribot's web app, answering questions about multiple companies (Fig.~\ref{fig:bose_third} to Fig.~\ref{fig:oyoty_double}). Next, we show screenshots from our web application for navigating the results produced by \framework (Fig.~\ref{fig:medium_first} to Fig.~\ref{fig:medium_choices}). These apps are available at \siteurl.

\begin{figure}[h]
\centering
\begin{minipage}{0.9\columnwidth}
\centering
  \includegraphics[width=1\linewidth]{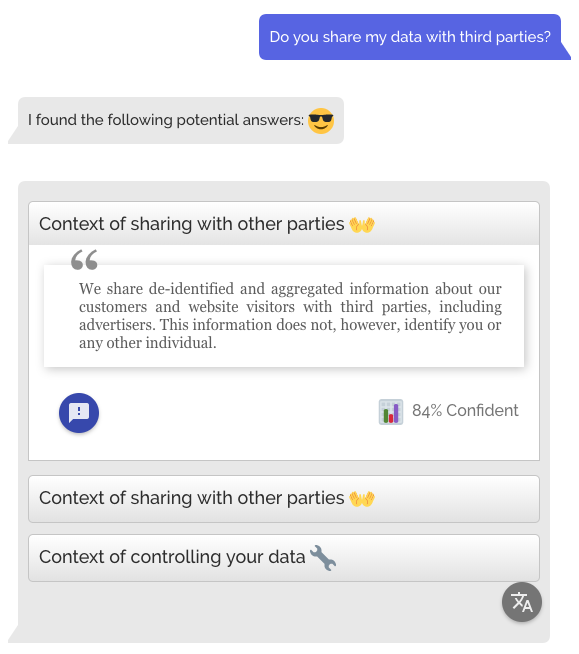}
  \caption{The first answer from our chatbot implementation of \pribot about third-party sharing in the case of Bose.com. Answers are annotated by a header mentioning the high level category (e.g., Context of sharing with third parties). The confidence metric is also highlighted int the interface.}
  \vspace{1\baselineskip}
\label{fig:bose_third}
\end{minipage}
\quad
\end{figure}

\begin{figure}[h]
\begin{minipage}{0.9\columnwidth}
\centering
  \includegraphics[width=1\linewidth]{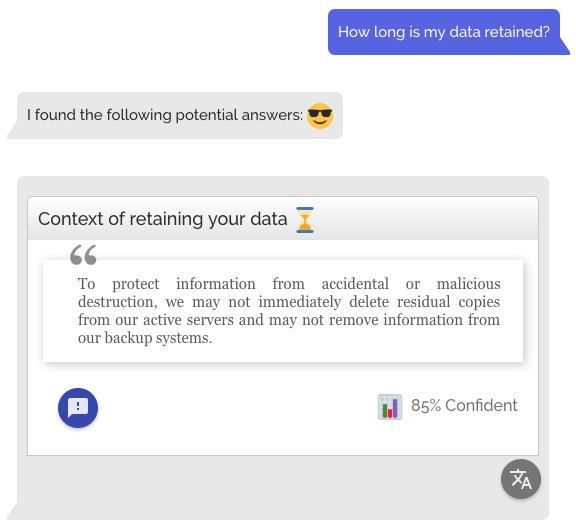}
  \caption{The first answer about data retention in the case of Medium. Notice the semantic matching in the absence of common terms. Notice also that only one answer is shown as it is the only one with high confidence. Hence, the user is not distracted by irrelevant answers.}
\label{fig:medium_retention}
\end{minipage}
\quad
\begin{minipage}{0.9\columnwidth}
\centering
  \includegraphics[width=1\linewidth]{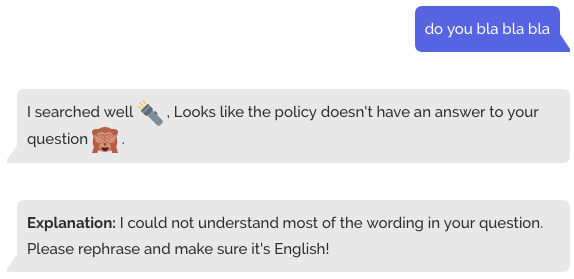}
  \caption{The answer given a question \textit{``do you bla bla bla''}, showcasing the power of the confidence metric, accounting for unknown words in the question and relaying that particular reason to the user.}
  \vspace{1\baselineskip}
\label{fig:wrong_english_oyoty}
\end{minipage}
\begin{minipage}{0.9\columnwidth}
\centering
  \includegraphics[width=1\linewidth]{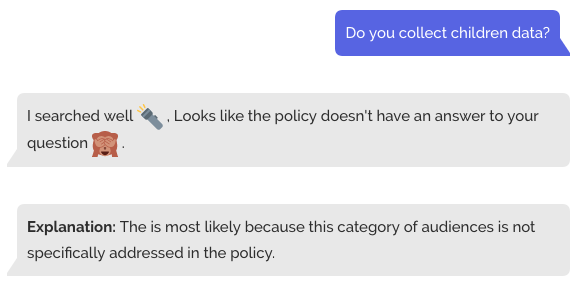}
  \caption{This case illustrates the scenario when \pribot finds no answer in the policy and explains the reason based on the automatically detected high-level category (explanations are preset in the application).}
\label{fig:twitter_children}
\end{minipage}
\end{figure}

\begin{figure}[h]
\begin{minipage}{0.9\columnwidth}
\centering
  \includegraphics[width=1\linewidth]{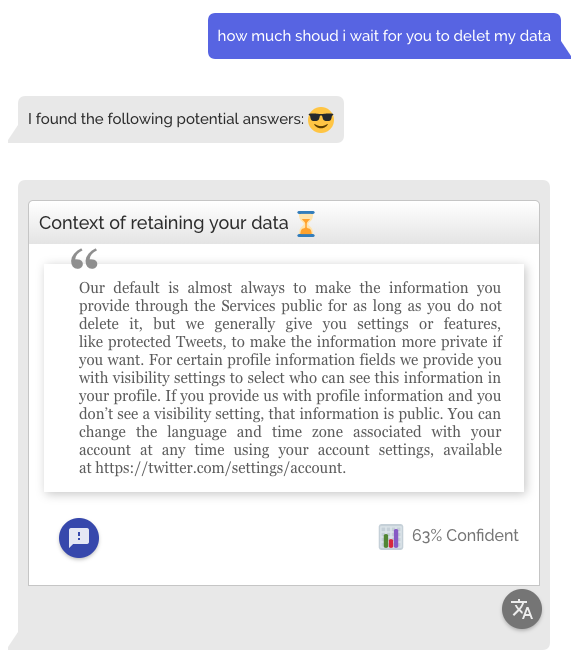}
  \caption{This case illustrates the power of subword embeddings. Given a significantly misspelled question \textit{``how much shoud i wait for you to delet my data''}, \pribot still finds the most relevant answer. }
\vspace{1\baselineskip}
\label{fig:mispelling_twitter}
\end{minipage}
\begin{minipage}{0.9\columnwidth}
\centering
  \includegraphics[width=1\linewidth]{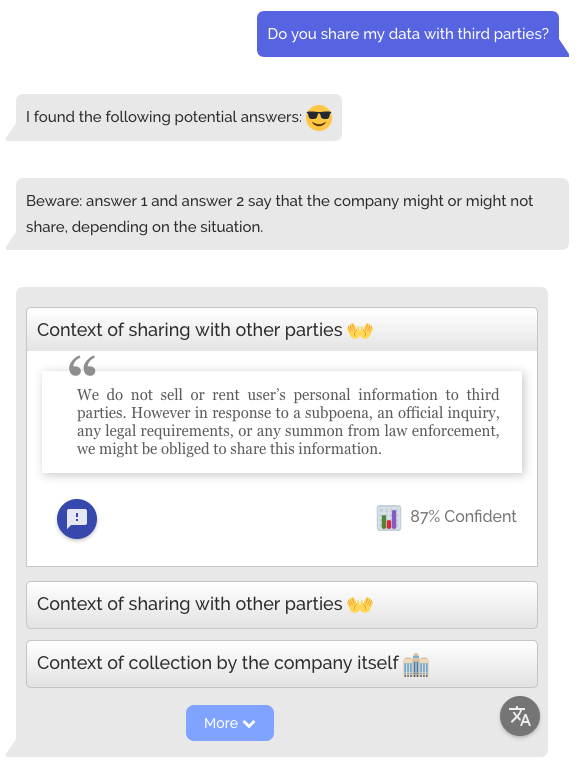}
  \caption{This case, with the policy of Oyoty.com, illustrates automatic accounting for discrepancies across segments (Sec.~\ref{sec:terms}) by warning the user about that.}
\label{fig:oyoty_double}
\end{minipage}
\end{figure}

\begin{figure*}[h]
\centering
\begin{minipage}{0.9\textwidth}
  \includegraphics[width=1\linewidth]{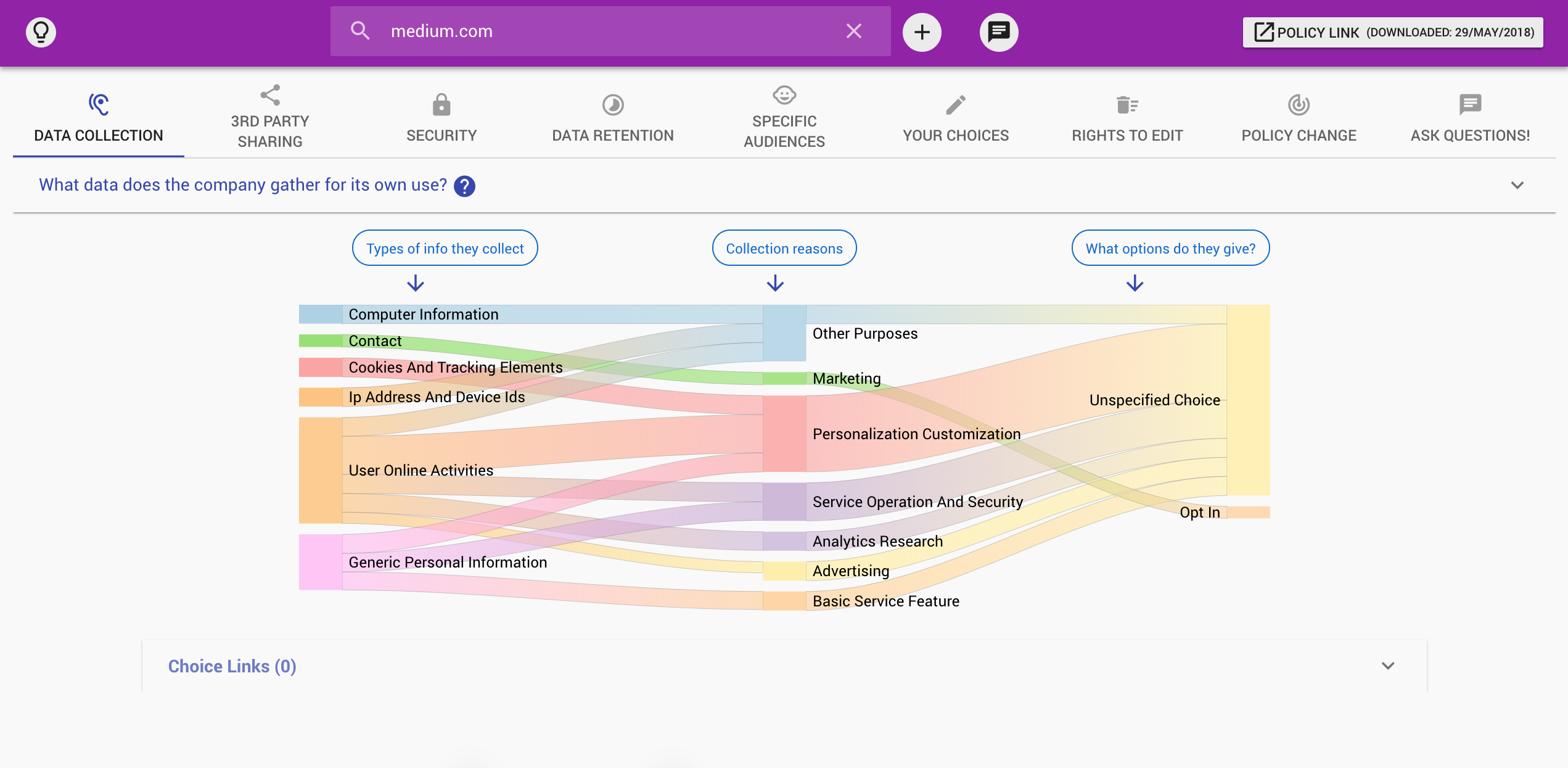}
  \caption{We show a case where our web app visualizes the result produced by \framework. The app shows the flow of the data being collected, the reasons behind that, and the choices given to the user in the privacy policy. The user can check the policy statements for each link by hovering over it.}
  \vspace{2\baselineskip}
\label{fig:medium_first}
\end{minipage}
\begin{minipage}{0.9\textwidth}
\centering
  \includegraphics[width=1\linewidth]{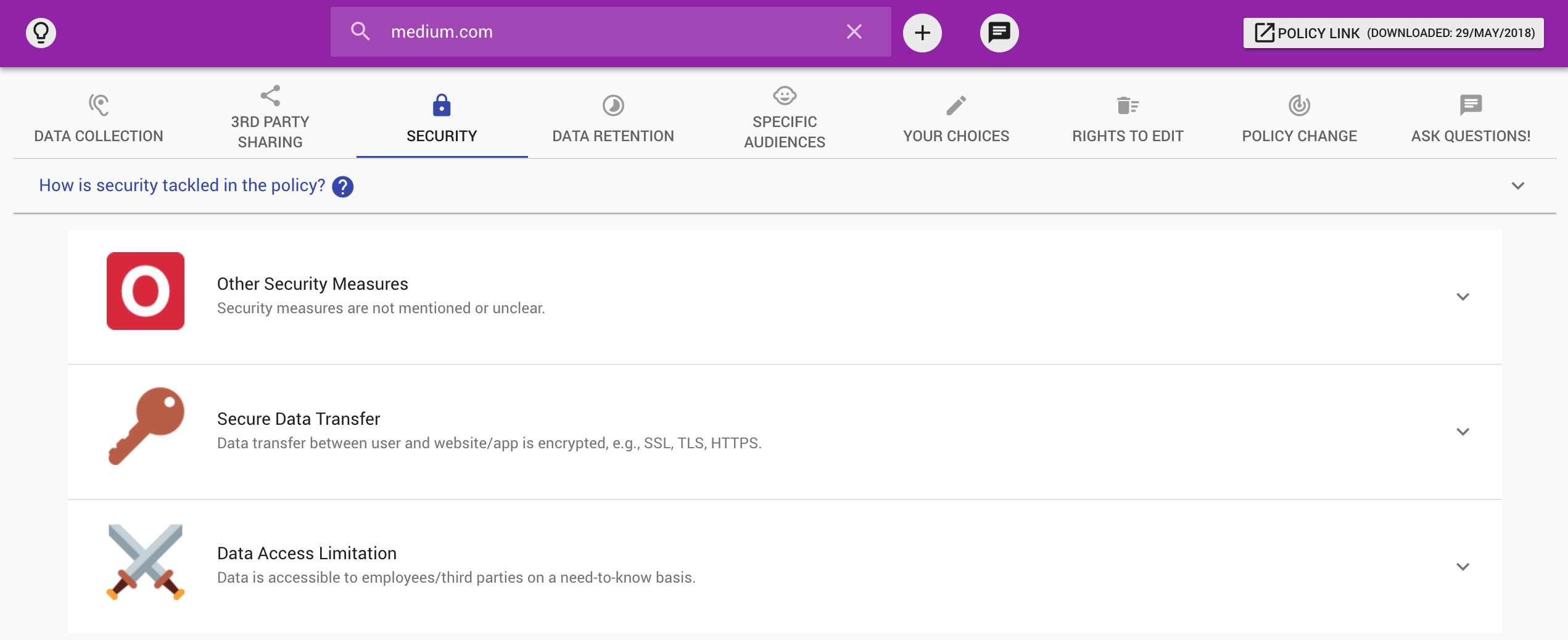}
  \caption{In this case, the security aspects of the policy are highlighted based on the labels extracted from \framework. The user has the option to see the related statement by expanding each item in the app.}
\label{fig:medium_security}
\end{minipage}
\quad

\end{figure*}

\begin{figure*}[h]
\centering
\begin{minipage}{0.9\textwidth}
  \includegraphics[width=1\linewidth]{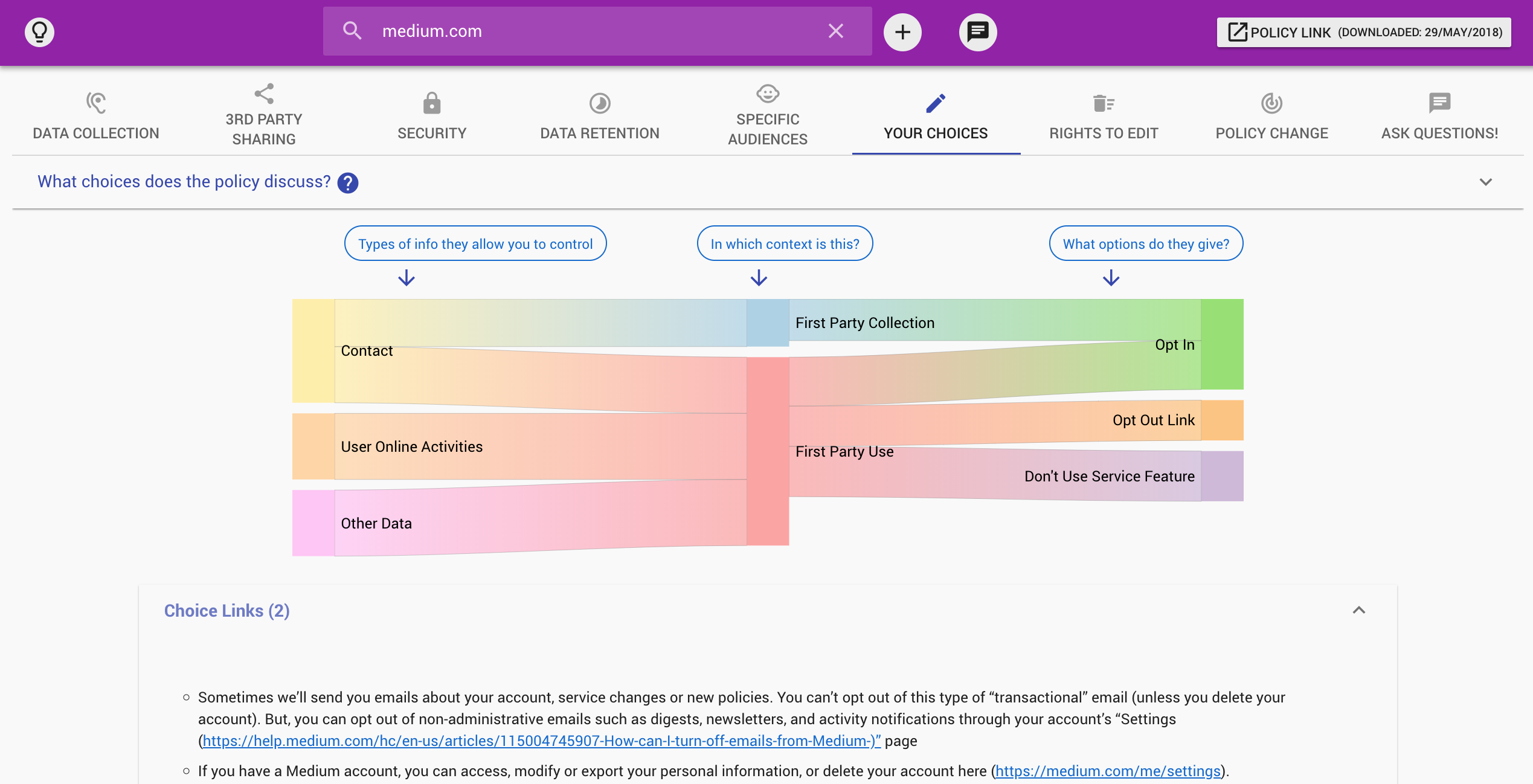}
  \caption{Here, the user is presented with the choices possible, automatically retrieved from \framework.}
\label{fig:medium_choices}
\vspace{2\baselineskip}
\end{minipage}
\end{figure*}

 \fi

\end{document}